\PassOptionsToPackage{table}{xcolor}
\documentclass{article} 
\usepackage[OT1]{fontenc}
\usepackage{iclr2027_conference,times}

\usepackage{amsmath,amsfonts,bm}

\def\eqref#1{equation~\ref{#1}}

\def\1{\bm{1}}

\DeclareMathAlphabet{\mathsfit}{\encodingdefault}{\sfdefault}{m}{sl}
\SetMathAlphabet{\mathsfit}{bold}{\encodingdefault}{\sfdefault}{bx}{n}

\usepackage{graphicx}
\usepackage{booktabs}
\usepackage{multirow}
\usepackage{array}
\usepackage[table]{xcolor}
\usepackage{xspace}
\usepackage{float}
\usepackage{tcolorbox}
\tcbuselibrary{breakable}
\usepackage{hyperref}
\usepackage{url}
\usepackage{adjustbox}

\usepackage{pgfplots}
\pgfplotsset{compat=1.18}
\usepackage{xcolor}
\usepackage{booktabs}
\usepackage{graphicx}
\usepackage{subcaption}
\definecolor{ScoreBlue}{HTML}{FDF4D2}
\definecolor{SuccessOchre}{HTML}{B0CDE6}
\definecolor{StepsSage}{HTML}{A290B7}
\definecolor{TokensPlum}{HTML}{946D6D}
\definecolor{GridGray}{HTML}{D9DDE2}
\usepackage{wrapfig}
\usepackage{needspace}
\usepackage{enumitem}

\newtcolorbox{promptbox}[1]{
    breakable,
    colback=gray!8,
    colbacktitle=gray!8,
    coltitle=black,
    boxrule=0pt,
    sharp corners,
    boxsep=0pt,
    left=12pt, right=12pt, top=8pt, bottom=12pt,
    toptitle=12pt, bottomtitle=0pt,
    before skip=8pt, after skip=8pt,
    fonttitle=\normalsize\sffamily\bfseries,
    fontupper=\small\rmfamily,
    before upper={\raggedright\setlength{\parskip}{5pt}\setlength{\parindent}{0pt}},
    lines before break=4,
    title={#1},
    title after break={#1\space(continued)}
}
\newlist{promptitems}{itemize}{2}
\setlist[promptitems]{
    label=\textbullet, leftmargin=1.4em, labelsep=0.5em,
    topsep=4pt, partopsep=0pt, itemsep=4pt, parsep=0pt,
    beginpenalty=10000
}
\setlist[promptitems,2]{label=\textendash,itemsep=3pt}
\newlist{promptsteps}{enumerate}{1}
\setlist[promptsteps]{
    label=\arabic*., leftmargin=1.7em, labelsep=0.5em,
    topsep=4pt, partopsep=0pt, itemsep=4pt, parsep=0pt,
    beginpenalty=10000
}
\makeatletter
\newcommand{\promptheading}[1]{%
    \par\addvspace{8pt}%
    {\sffamily\bfseries #1\par}%
    \nobreak\@afterheading
}
\makeatother
\newenvironment{promptcode}
    {\par\begingroup\ttfamily\raggedright\setlength{\parskip}{0pt}}
    {\par\endgroup}

\usepackage{tikz}
\usepackage{xcolor}
\usepackage{graphicx}
\usepackage{capt-of}

\definecolor{ScoreVeryLow}{HTML}{F3F5F7}
\definecolor{ScoreLow}{HTML}{DDE5ED}
\definecolor{ScoreMid}{HTML}{B0C1D1}
\definecolor{ScoreHigh}{HTML}{819BB4}

\definecolor{RateVeryLow}{HTML}{F7F3ED}
\definecolor{RateLow}{HTML}{EBDCCB}
\definecolor{RateMid}{HTML}{DAB793}
\definecolor{RateHigh}{HTML}{C18C5D}

\definecolor{deltared}{RGB}{200, 50, 50}
\definecolor{bestrow}{RGB}{248, 255, 235}
\definecolor{deltagreen}{RGB}{34, 139, 34}

\usepackage{tabularx}

\title{Work2Skill: Capturing Human GUI Experience for Reusable Agent Skill Induction}

\author{%
\adjustbox{max width=\dimexpr\textwidth-2\tabcolsep\relax}{%
\textbf{Yuqi Huang}\textsuperscript{\normalfont \(1,\ddag\)}\quad
\textbf{Hui Li}\textsuperscript{\normalfont \(2,\ddag\)}\quad
\textbf{Yifei Cao}\textsuperscript{\normalfont 2}\quad
\textbf{Zikun Cai}\textsuperscript{\normalfont 3}\quad
\textbf{Tianci Xie}\textsuperscript{\normalfont 4}\quad
\textbf{Yangfan Gao}\textsuperscript{\normalfont 2}} \\[3pt]
\adjustbox{max width=\dimexpr\textwidth-2\tabcolsep\relax}{%
\textbf{Pinxue Guo}\textsuperscript{\normalfont \(5,\dagger\)}\quad
\textbf{Chaoyun Zhang}\textsuperscript{\normalfont \(5,\dagger\)}\quad
\textbf{Bingxuan Li}\textsuperscript{\normalfont 5}\quad
\textbf{Rui Zheng}\textsuperscript{\normalfont 5}\quad
\textbf{Hang Yan}\textsuperscript{\normalfont 5}\quad
\textbf{Ning Liao}\textsuperscript{\normalfont \(1,*\)}} \\[8pt]
\adjustbox{max width=\dimexpr\textwidth-2\tabcolsep\relax}{%
\normalfont\footnotesize
\textsuperscript{1}Shanghai Jiao Tong University\quad
\textsuperscript{2}Fudan University\quad
\textsuperscript{3}Nanyang Technological University} \\[2pt]
\adjustbox{max width=\dimexpr\textwidth-2\tabcolsep\relax}{%
\normalfont\footnotesize
\textsuperscript{4}Zhejiang University\quad
\textsuperscript{5}Shanghai Qiji Zhifeng Co., Ltd} \\[5pt]
\normalfont\footnotesize
\textsuperscript{\(\ddag\)}Equal contribution\quad
\textsuperscript{\(\dagger\)}Project Leaders\quad
\textsuperscript{*}Corresponding Author
}

\newcommand{\ours}{Work2Skill\xspace}

\DeclareRobustCommand{\toolname}[1]{\texttt{#1}}

\iclrfinalcopy 
\begin{document}

\maketitle
\fancyhead{}
\fancyhead[L]{Preprint}

\begin{abstract}
Computer-use agents can perform increasingly complex GUI tasks, yet unfamiliar
software and workflows requiring human domain knowledge remain challenging.
Human demonstrations convey procedural knowledge through the same GUI
interaction space, but reuse requires interpreting actions in context and
separating workflow rules from task-specific details. We thus introduce
Work2Skill, a framework that captures human GUI
experience for reusable agent skill induction. WorkRecorder provides the
framework's infrastructure for collecting this experience, aligning screen
recordings, mouse and keyboard events, and available accessibility-tree
information. Using WorkRecorder, we
construct WorkDemo-350, the first synchronized multimodal dataset of human
desktop GUI demonstrations for reusable agent skill induction and downstream
evaluation. Its 350 demonstrations are screened for completeness and task
success.
We then propose an agentic skill-induction method that inspects evidence on
demand to produce reusable skills capturing workflow rules, step dependencies,
and completion checks.
When the same four execution models are equipped with skills induced by each
method under shared execution settings, Work2Skill achieves the highest scores
and success rates.
Mean score improvements over the corresponding No Skill baseline are
\(4.67\)--\(10.54\) points on MyPCBench Bounded Action and
\(3.11\)--\(7.50\) points on a 286-task subset of OSWorld-Verified.
Our code and data are available at
\url{https://github.com/china-qijizhifeng/Work2Skill}.
\end{abstract}

\section{Introduction}
\begin{wrapfigure}[18]{r}{0.3\textwidth}
    \vspace{-\intextsep}
    \centering
    \includegraphics[width=\linewidth]{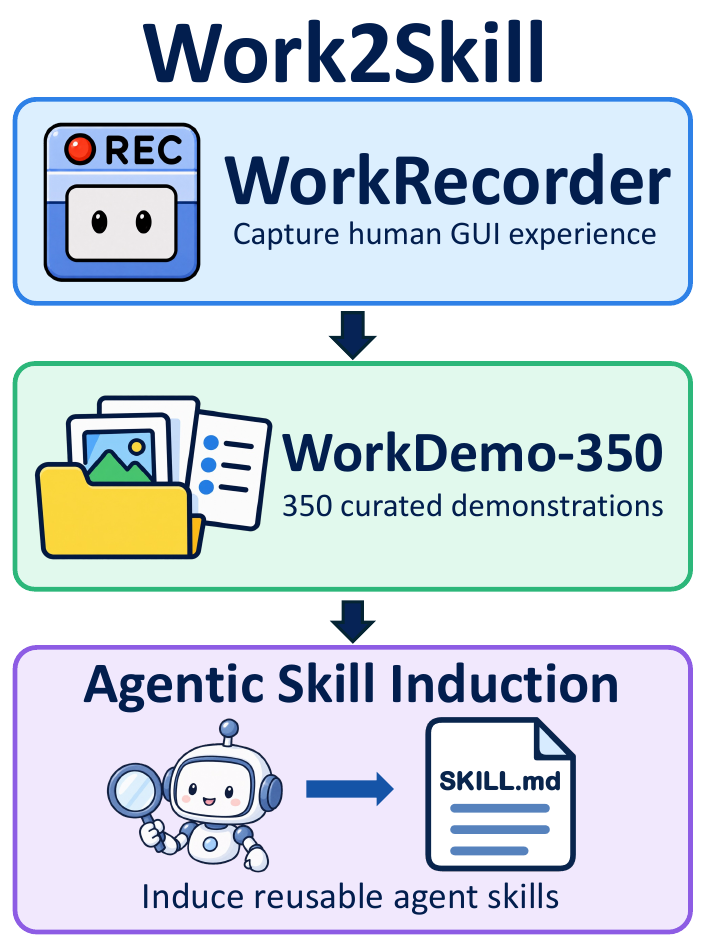}
    \captionsetup{font=footnotesize,skip=3pt,justification=raggedright,singlelinecheck=false}
    \caption{Work2Skill framework: WorkRecorder, WorkDemo-350,
    and Agentic Skill Induction.}
    \label{fig:work2skill-intro}
    \vspace{-\intextsep}
\end{wrapfigure}
Computer-use agents can perform increasingly complex desktop and web tasks,
supported by advances in GUI grounding and planning
\citep{qin2025uitars,agashe2025agents2,wang2025opencua}.
Yet agents still struggle with unfamiliar software and workflows that require
domain knowledge, where successful execution depends on procedures not spelled
out in the user's request \citep{zhang2026showuialoha,jang2026mypcbench}.
For example, ``submit my reimbursement'' does not specify which internal portal
to use, which receipts to attach, or what to check before submission. Agents
need this user- or organization-specific procedural knowledge to translate a
user's goal into GUI actions.

Human demonstrations offer a direct source of this knowledge because humans and
computer-use agents share the same GUI interaction space. Demonstrated controls,
action sequences, and visible outcomes can therefore guide an agent's own
execution.
Prior work converts demonstrations into expert instructions and semantic action
traces \citep{li2025instructionagent,zhang2026showuialoha}, multimodal tutorials
\citep{bai2026demo2tutorial}, or reusable skills
\mbox{\citep{yang2026browserbc,xiao2026omegause}}.
OpenAI's Record \& Replay \citep{openai2026recordreplay} and Claude's Record a
skill \citep{anthropic2026recordskill} bring this approach to user-facing tools.
We thus introduce \ours, a framework that captures human GUI experience for
reusable agent skill induction (Figure~\ref{fig:work2skill-intro}). Its three
components address the following challenges in action grounding, method
comparison, and evidence inspection.

First, action coordinates reveal where an operation occurred but not what the target
control means. ShowUI-Aloha interprets actions from full-screen frames and target
crops \citep{zhang2026showuialoha}, while Demo2Tutorial uses before-and-after
screenshots \citep{bai2026demo2tutorial}. Without accessibility-tree (AX)
evidence, these parsers depend on visual grounding to infer target semantics;
selected frames can also miss transient states and intermediate operations.
WorkRecorder addresses this ambiguity by aligning screen recordings,
mouse and keyboard events, and available AX information. Recorded element names
and roles provide explicit control semantics, while linked video preserves the
surrounding state changes. The induction agent can thus interpret an operation
using both its visual context and available structured UI evidence.

Second, representative desktop GUI studies use different demonstrations, task
sets, and execution protocols \citep{zhang2026showuialoha,bai2026demo2tutorial}.
Their results mix differences in induction with differences in source
experience and evaluation conditions. WorkDemo-350 provides a common
demonstration resource for skill induction in desktop environments.
Collected with WorkRecorder, it contains 350
human GUI demonstrations covering 64 MyPCBench Bounded Action tasks
\citep{jang2026mypcbench} and a 286-task OSWorld-Verified subset
\citep{xie2024osworld}. Each demonstration is screened for completeness and task
success and packaged with synchronized visual and action evidence, plus
available AX information. This shared evidence and a controlled downstream protocol support comparisons of
induction methods using the same task evaluators and execution conditions.

Third, representative pipelines prescribe an order for action parsing and procedure
abstraction, even when individual stages refine their outputs
\citep{zhang2026showuialoha,bai2026demo2tutorial}. This couples evidence inspection
to predefined stages. Loading a long recording into context at once also creates
context pressure, while selecting a fixed subset can omit details needed later.
Our agentic skill-induction method addresses both constraints by
interleaving on-demand evidence inspection with skill writing and revision.
Drawing on active evidence acquisition in agentic video understanding
\citep{yuan2025videoexplorer,wang2025avp,li2026lenswalk}, the agent selects linked
files, images, and video intervals to resolve questions as they arise. It can
revisit the source after drafting without loading the entire recording at once.
During skill writing, we guide the agent to generalize demonstration-specific
inputs into task-dependent descriptions while preserving procedural rules,
step dependencies, and completion checks. The skills guide execution without
the source recording.

We evaluate the resulting skills through downstream execution on the
WorkDemo-350 tasks. Using official evaluators and a shared text-guidance
interface, we compare \ours with No Skill and text-adapted versions of
ShowUI-Aloha and Demo2Tutorial. Across four execution models, \ours improves
mean evaluation scores over No Skill by \(4.67\)--\(10.54\) points on MyPCBench
Bounded Action and \(3.11\)--\(7.50\) points on the evaluated OSWorld-Verified
subset. Further experiments test reuse under changed requirements within the
same application (MyPCBench Jitter) and transfer between two execution models.
Evidence ablations and the Pipeline Prompt comparison examine the roles of
input modalities and induction-prompt design. Our contributions are threefold:

1) We introduce \textbf{WorkRecorder}, the human GUI experience collection infrastructure
of \ours, which aligns recorded inputs with visual and available AX evidence
to clarify demonstrated actions.

2) We construct \textbf{WorkDemo-350}, the first synchronized multimodal dataset
of human desktop GUI demonstrations for reusable agent skill induction and
downstream evaluation.

3) We propose an \textbf{agentic skill induction method} that interleaves on-demand
evidence inspection with skill writing and revision. Across two benchmarks and
four execution models, it achieves the highest scores and success rates among
the evaluated methods.

\section{Related Work}
\label{sec:related}

\paragraph{Learning from Human GUI Demonstrations.}
Human demonstrations provide experience for GUI agents across desktop, web,
and mobile environments
\citep{nguyen2025guiagentsurvey,tang2025guiagentsurvey,qin2025uitars}.
They support data collection, grounding, model training
\citep{wang2025opencua,nayak2025uivision,feizi2025groundcua} and agent
efficiency evaluation \citep{abhyankar2025osworldhuman}. Demonstration structure
can also influence task execution
\citep{xing2026hierarchical}.
Procedural representations include expert instructions \citep{li2025instructionagent},
task graphs \citep{yin2025taskmind}, and parameterized workflows for robust replay
\citep{zhao2026gpa}. LearnAct also introduces LearnGUI for evaluating
demonstration-based learning on mobile devices \citep{liu2025learnact}.
More recent work includes semantic
action traces, multimodal tutorials, natural-language skill graphs, and standard
operating procedure (SOP) skills
\citep{zhang2026showuialoha,bai2026demo2tutorial,yang2026browserbc,xiao2026omegause}.

\paragraph{Skill Induction and Active Evidence Acquisition.}
\label{sec:related-skills}

Reusable guidance draws on experience organization and evidence selection.
Skill-induction methods learn procedures through autonomous
interaction \citep{wang2025asi,zheng2025skillweaver} or distill execution and user
traces \citep{mi2026skillpro,ni2026trace2skill,yang2026autoskill}, with environment
feedback and policy learning supporting further refinement
\citep{xia2026skillrl,yang2026skillforge}. Adaptive memories, contextual
playbooks, and learned memory-management routines retain experience
\citep{suzgun2025dynamiccheatsheet,zhang2025ace,zhang2026memskill},
while experience-based evaluation examines its downstream value
\citep{peng2026evoclawbench}.
Guidance from multimodal demonstrations requires choosing what to inspect.
Agentic video methods retrieve and examine relevant
segments iteratively
\citep{ma2025drvideo,yang2025vca,yuan2025videoexplorer,wang2025avp,li2026lenswalk}.
In GUI settings, CUADebug inspects traces and screenshots to diagnose execution
failures \citep{zhang2026cuadebug}, while MMSkills packages textual procedures,
state cards, and keyframes for selective inspection during execution
\citep{zhang2026mmskills}. 

\section{Work2Skill Framework}
\label{sec:framework}

\begin{figure}[!t]
\centering
\includegraphics[width=\linewidth]{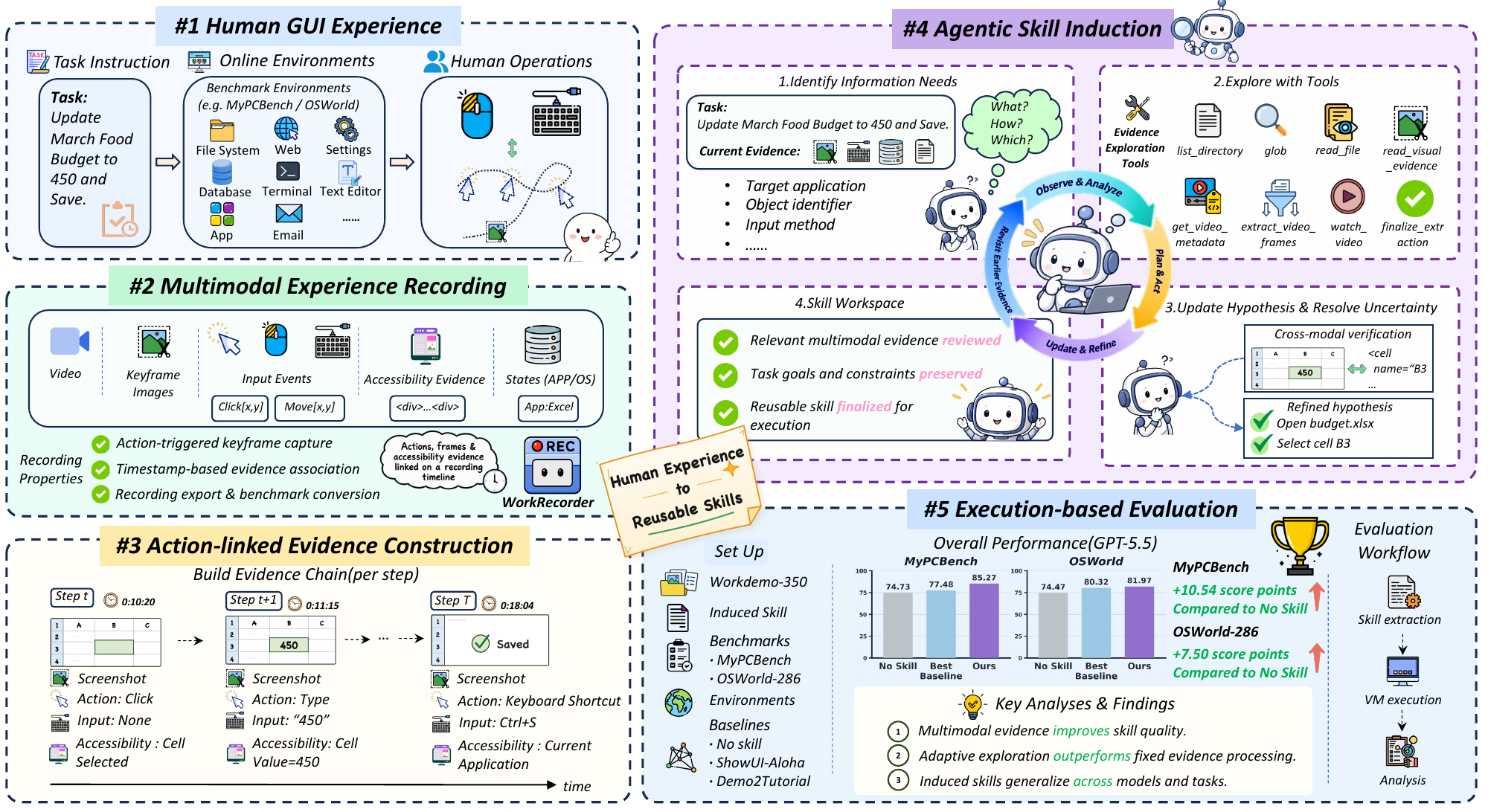}
\caption{Overview of Work2Skill. WorkRecorder aligns human GUI actions with
visual and available accessibility evidence to build WorkDemo-350. An agent
inspects evidence on demand and revises reusable skills for downstream GUI
execution. Downstream evaluation compares these skills with guidance from
the fixed workflows of ShowUI-Aloha and Demo2Tutorial.}
\vspace{-15pt}
\label{fig:work2skill-overview}
\end{figure}

\ours connects human demonstrations to reusable procedural guidance
(Figure~\ref{fig:work2skill-overview}). WorkRecorder preserves actions together
with their visual and semantic context; WorkDemo-350 organizes the collected
demonstrations into a curated evidence resource. An induction agent inspects
this evidence, revisits unclear steps, and writes skills for downstream GUI
execution.

\subsection{WorkRecorder}
\label{sec:recorder}
\begin{figure}[t]
    \centering
    \includegraphics[width=1.0\linewidth]{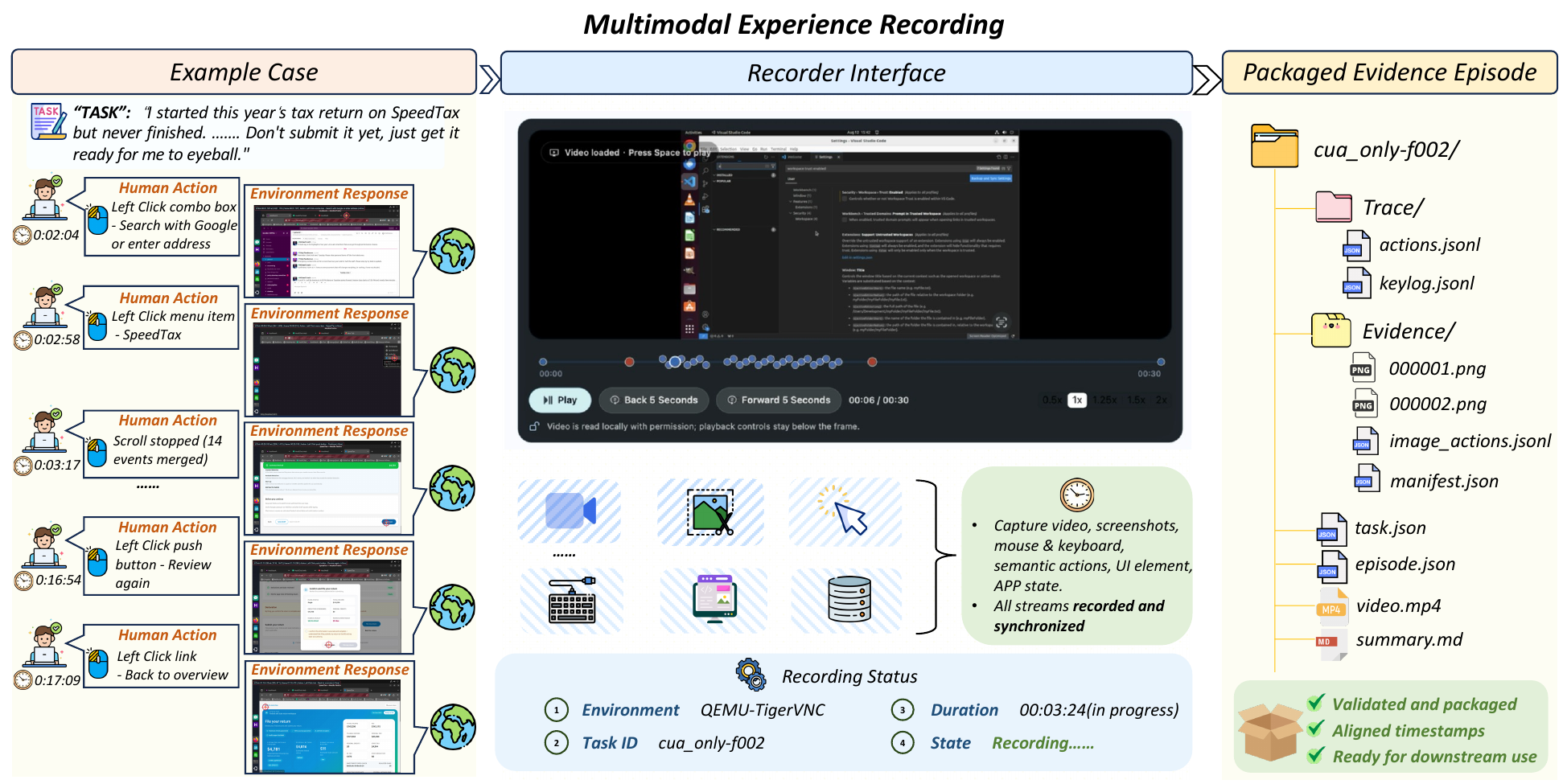}
    \caption{Human GUI experience collection with WorkRecorder. A demonstrated workflow (left) is captured and reviewed through the recorder interface (center), then organized into an evidence episode with task metadata, action traces, and available multimodal evidence (right).}
    \label{fig:recorder}
    \vspace{-12pt}
\end{figure}

WorkRecorder collects the evidence needed to interpret human actions in context.
Given a task instruction, a human demonstrator
performs the task in the benchmark's official environment. WorkRecorder captures
screen video, keyframe images, mouse and keyboard activity, and available UI
information as multimodal evidence (Figure~\ref{fig:recorder}). An action
trace links each recorded operation to its supporting evidence so that the
induction agent can inspect individual steps. Task setup and outcome evaluation
follow the corresponding benchmark procedures.

A macOS interface records local workflows or interactions with a remote Linux
desktop. For remote tasks, the Mac captures the displayed workflow and local
inputs, while a collector in the target environment supplies UI evidence.
Platform-specific collectors provide target-element metadata, accessibility-tree
(AX) snapshots, and application or window context when available. Coverage
depends on the recording configuration and the application's accessibility
support.

\paragraph{Multimodal Capture and Alignment. }
\label{sec:capture-alignment}
WorkRecorder aligns input events and visual observations on a common
video-relative timeline. It groups timestamped events into actions such as
clicks, drags, scrolling, and hotkeys, then links these actions to keyframes.
An agent can therefore inspect a click alongside its screen image, available
target information, and surrounding video to determine what changed.
Target-element association uses pointer location and event timing, with window
context and control semantics helping select among candidates. On Linux,
AT-SPI point queries and nearby accessibility events supply available names,
roles, and bounds for click and drag records. UI annotations are derived from
the corresponding target elements in the AX tree. AX snapshots provide
additional interface context but need not reflect the exact action-time state.
Appendix~\ref{app:recorder} details the recording architecture, capture policies,
target-matching rules, and time alignment.

\paragraph{Evidence Representation and Packaging. }
\label{sec:evidence-representation}
Each demonstration becomes an evidence episode containing the task
instruction and metadata, an ordered action trace, and the available multimodal
evidence. A manifest indexes these resources; action records link to keyframes
and video times. These links support targeted inspection without processing
the recording sequentially. Descriptions help locate operations; linked
recordings let the agent verify their interpretation. The
available evidence depends on capture and export settings.
Appendix~\ref{app:schema} provides details.

\subsection{WorkDemo-350}
\label{sec:collection-stats}
WorkDemo-350 provides a common evidence resource for studying skill induction.
Collected with WorkRecorder, it contains 350 human GUI demonstrations: 64
MyPCBench Bounded Action tasks and a fixed 286-task OSWorld-Verified subset.
The two benchmarks provide complementary settings. MyPCBench
\citep{jang2026mypcbench} covers personal office and everyday workflows with
user-specific context; its Bounded Action tasks offer scoped GUI workflows
suited to human demonstration. OSWorld-Verified \citep{xie2024osworld}
adds general computer-use tasks across desktop and web applications, broadening
the range of workflows available for skill induction and evaluation.

\begin{wraptable}[13]{r}{0.58\linewidth}
    \vspace{-8pt}
    \centering
    \small

    \caption{Statistics of WorkDemo-350: 350 human GUI demonstrations covering
    64 MyPCBench Bounded Action tasks and a 286-task subset of
    OSWorld-Verified, with aligned
    multimodal evidence.}
    \label{tab:data-stats}
    \vspace{-5pt}
    \setlength{\tabcolsep}{3pt}
    \renewcommand{\arraystretch}{1.10}

    \begin{tabularx}{\linewidth}{@{}Xrr@{}}
        \toprule
        Statistic
        & MyPCBench
        & OSWorld-Verified \\
        \midrule

        Number of tasks             & 64      & 286      \\
        Total video duration (s)     & 7623.02 & 22890.18 \\
        Mean episode duration (s)    & 119.11  & 80.04    \\
        Mean actions per episode    & 32.23   & 34.57    \\
        Mean evidence images        & 24.20   & 23.91    \\

        \bottomrule
    \end{tabularx}
\end{wraptable}

We screen tasks and recordings to retain successful, high-quality demonstrations.
For OSWorld-Verified, we first exclude infeasible tasks for which a valid human
demonstration cannot be completed. Across both benchmarks, demonstrators follow
valid GUI workflows and verify the final state without being restricted to a
prescribed action sequence. We then review task success and recording
completeness, discarding demonstrations that omit key operations or end in an
incorrect state (Appendix~\ref{app:benchmark}). Each episode links task and
action records to available visual and UI evidence.

Episodes are organized by benchmark and task so that methods can draw on the
same demonstrations while choosing which modalities and time intervals to
inspect. The collection contains approximately 8.48 hours of recorded
experience, accompanied by action traces and evidence images.
Table~\ref{tab:data-stats} summarizes the dataset's size and multimodal evidence
statistics.

\subsection{Agentic Skill Induction}
\label{sec:induction}

Skill induction turns an evidence episode into reusable procedural guidance.
Given the task instruction and recorded evidence, a NexAU tool-using agent
\citep{nexagi2026nexau} generates skills in an output directory separate from the
recordings. The default system prompt is concise: it sets evidence-access
boundaries, prerequisite checks, and completion requirements while leaving the
detailed exploration and revision sequence to the agent. A bundled
\texttt{skill-creator} policy guides skill authoring. After reading this policy
and the task definition in \texttt{task.json}, the agent alternates evidence
inspection with skill writing and revision. Appendix~\ref{app:prompts} provides
the prompts and authoring policy.

\paragraph{Evidence Exploration Tools. }
\label{sec:evidence-tools}
We select file, image, and video tools that expose complementary views of a
demonstration at different granularities. File discovery and reading provide
task definitions, manifests, action traces, key logs, and available structured
UI evidence. Traces and indices link operations to their supporting media,
while AX records supply available control names and roles.
\toolname{read\_visual\_evidence} returns selected images with source-path
labels, exposing visible controls and entered or selected values. Video tools
provide access to states and transitions beyond the indexed images. After
reading video metadata, the agent can sample an interval with
\toolname{watch\_video} or request individual timestamps with
\toolname{extract\_video\_frames}; both return timestamp-labeled frames.
Source-path and timestamp labels preserve the correspondence between returned
observations and source artifacts. Appendix~\ref{app:tools} details the
interfaces, and Section~\ref{sec:induction-ablation} evaluates the contribution
of the available evidence to downstream execution.

\paragraph{Active Evidence Acquisition. }
\label{sec:active-acquisition}
The agent chooses its next observation according to the task and unresolved
questions in the evidence, following iterative inspection in agentic video
understanding \citep{yang2025vca,wang2025avp,li2026lenswalk}. The prompt requires
representative initial, critical, and final-state images to be inspected when
indexed images are available; the agent otherwise determines the inspection
sequence and level of detail. For example, an ambiguous click can prompt
inspection of the target frame to identify the control, followed by a later
video frame to check the resulting state. Tool guidance encourages coarse
video sampling followed by closer examination of relevant actions.
Evidence inspection and skill writing share one conversation, so a draft can
expose missing constraints or unverified transitions.
The agent can then return to the relevant evidence and revise the procedure
to incorporate supported dependencies and completion checks. Candidate skills
are not executed during induction, and downstream evaluation results are not
used as revision feedback. When the
available observations do not resolve an uncertainty, the authoring policy
requires an observable conditional check or an explicit limitation in the
skill. Section~\ref{sec:induction-ablation} compares this prompt with Pipeline
Prompt, which sets exploration order but allows further inspection.

\paragraph{Skill Generation. }
\label{sec:skill-generation}
The agent expresses each coherent reusable capability in a \texttt{SKILL.md}, with
YAML frontmatter containing its name and description, following the Agent
Skills file convention \citep{zhang2025agentskills}. Through the authoring
policy, we guide the agent to generalize demonstration-specific inputs into
task-dependent descriptions and describe UI targets semantically. The policy
instructs the agent to keep phases together when they share state, task-derived
data, quantities, or required ordering, and to retain selection rules,
calculations, dependencies, requested scope, stop conditions, visible completion
checks, and supported recovery. Skills must preserve visible GUI interaction
and include all instructions needed for execution without the source recording.
Before completion, the agent calls \toolname{finalize\_extraction} to validate
output structure and required tool use. Checks cover task-file access, required
skill structure, a recorded write followed by a read of each final skill,
and metadata inspection before video sampling. Successful finalization
records skill hashes, inspected source paths, validation results, and run
provenance. These checks verify output structure and required tool use;
they do not verify skill correctness or task success.
For downstream evaluation, the executor receives the target task instruction
and generated skill. It selects GUI actions using this guidance together with
its observations and action history, without accessing the source recording.
Sections~\ref{sec:cross-model} and
\ref{sec:task-generalization} evaluate reuse across execution models and task
instances. Appendix~\ref{app:skill-output} details the authoring policy and
output requirements.

\section{Experiments}
\label{sec:experiments}

We organize our experiments around the following four research questions.
RQ1 tests whether the induced
skills improve execution on the demonstrated tasks. RQ2 and RQ3 then examine
two dimensions of reuse: changes to task requirements and changes to the
execution model, respectively. RQ4 investigates how the available evidence and
induction prompt affect downstream utility. A case study illustrates how
induced skills handle dependent subgoals in a multistep workflow.

\subsection{Experimental Setup}
\label{sec:experimental-setup}

\paragraph{Benchmarks.}
We evaluate on the 64 MyPCBench Bounded Action tasks and 286 OSWorld-Verified
tasks in WorkDemo-350 using official task evaluators. On MyPCBench, the
execution agent has access to both GUI actions and Bash commands. Benchmark
selection and demonstration screening are described in
Section~\ref{sec:collection-stats}. OSWorld results apply to this curated subset.
Appendix~\ref{app:evaluator-config} describes our OSWorld execution harness
and evaluator configuration.

\paragraph{Skill-based Evaluation.}
We evaluate skills as plug-in guidance by appending their text as a clearly
delimited block to the end of the executor's base system prompt. All
text-guidance methods use the same \texttt{<task\_skills>} wrapper;
Appendix~\ref{app:execution-prompt} gives the template and injected contents.
Separating procedural guidance
from the execution loop supports
reuse across execution models and integration with different agent harnesses.
We therefore use shared execution settings for each model and benchmark,
varying only the appended guidance to control for differences in native agent
pipelines. We compare \ours with No Skill, action semantics from
ShowUI-Aloha (Aloha) \citep{zhang2026showuialoha}, and tutorial text from
Demo2Tutorial (D2T) \citep{bai2026demo2tutorial}. The demonstration-based methods
use the same source episodes, while their input modalities and evidence
representations follow each method's design. RQ1 thus compares their overall
ability to produce useful guidance under shared execution conditions;
RQ4 separately examines evidence and induction-prompt choices within \ours
while holding the model and skill-authoring policy fixed.

\paragraph{Models and Metrics.}
We use GPT-5.5 \citep{openai2026gpt55}, Opus-4.8
\citep{anthropic2026opus48}, Qwen3.5-35B-A3B, and Qwen3.5-9B
\citep{qwen2026qwen35}. By default, the same model induces skills and executes
tasks with them; RQ3 varies this pairing. We report mean evaluation score (Score, 0--100),
full-credit success rate (SR), mean execution steps, and mean skill-induction
token usage. Appendix~\ref{app:benchmark} details task selection, metrics,
baseline adaptations, and execution settings.

\begin{table*}[t]
\centering
\caption{
Performance and efficiency on MyPCBench and the evaluated
OSWorld-Verified subset. Bold indicates the best Score or SR within
each model and benchmark, and blue shading denotes our method.
Score is the mean evaluation score (0--100), and SR is the percentage of
tasks receiving full credit on either benchmark. Steps denotes average execution
steps, and Tok.\ (k) denotes average token usage for skill induction, in
thousands. Aloha denotes ShowUI-Aloha.
}
\label{tab:main-results}

\small
\setlength{\tabcolsep}{2pt}
\renewcommand{\arraystretch}{1.15}

\begin{adjustbox}{max width=\textwidth,center}
\begin{tabular}{@{}l *{4}{cccc}@{}}
\toprule

\multirow{2}{*}{Method}
& \multicolumn{4}{c}{GPT-5.5}
& \multicolumn{4}{c}{Opus-4.8}
& \multicolumn{4}{c}{Qwen3.5-35B-A3B}
& \multicolumn{4}{c}{Qwen3.5-9B} \\

\cmidrule(lr){2-5}
\cmidrule(lr){6-9}
\cmidrule(lr){10-13}
\cmidrule(lr){14-17}

& Score & SR (\%) & Steps & Tok. (k)
& Score & SR (\%) & Steps & Tok. (k)
& Score & SR (\%) & Steps & Tok. (k)
& Score & SR (\%) & Steps & Tok. (k) \\

\midrule

\rowcolor{gray!12}
\multicolumn{17}{c}{\textit{MyPCBench ($N=64$)}} \\

No Skill
& 74.73 & 23.44 & 40.89 & --
& 81.53 & 42.19 & 40.58 & --
& 38.38 & 9.38  & 56.28 & --
& 24.27 & 6.25  & 70.92 & -- \\

Aloha
& 77.48 & 37.50 & 41.33 & 170.65
& 79.27 & 34.38 & 37.94 & 256.23
& 37.69 & 12.50 & 67.47 & 243.86
& 26.61 & 3.13  & 79.77 & 235.36 \\

D2T
& 75.25 & 37.50 & 46.13 & 177.32
& 81.28 & 42.19 & 36.84 & 234.73
& 40.61 & 9.38  & 63.45 & 211.98
& 27.75 & 6.25  & 70.84 & 180.48 \\

\rowcolor{blue!8}
\textbf{Ours}
& \textbf{85.27} & \textbf{48.44}
& 47.38 & 334.52
& \textbf{86.20} & \textbf{50.00}
& 36.03 & 300.21
& \textbf{44.56} & \textbf{20.31}
& 46.69 & 245.63
& \textbf{34.11} & \textbf{12.50}
& 69.30 & 249.87 \\

\midrule

\rowcolor{gray!12}
\multicolumn{17}{c}{\textit{OSWorld-Verified ($N=286$)}} \\

No Skill
& 74.47 & 70.98 & 19.05 & --
& 79.23 & 77.97 & 26.37 & --
& 32.52 & 32.17 & 66.61 & --
& 27.48 & 26.92 & 70.02 & -- \\

Aloha
& 72.57 & 69.93 & 21.61 & 146.98
& 78.91 & 76.57 & 18.04 & 132.24
& 31.08 & 30.42 & 58.01 & 369.27
& 24.81 & 24.13 & 72.51 & 319.98 \\

D2T
& 80.32 & 76.22 & 21.46 & 134.93
& 80.66 & 78.67 & 17.65 & 165.80
& 33.93 & 32.52 & 65.52 & 219.82
& 26.45 & 25.52 & 73.84 & 175.27 \\

\rowcolor{blue!8}
\textbf{Ours}
& \textbf{81.97} & \textbf{78.67}
& 25.78 & 185.19
& \textbf{82.34} & \textbf{79.37}
& 20.15 & 238.66
& \textbf{36.36} & \textbf{36.01}
& 65.89 & 218.88
& \textbf{33.41} & \textbf{32.52}
& 68.13 & 261.08 \\

\bottomrule
\end{tabular}
\end{adjustbox}

\vspace{-5pt}
\end{table*}

\subsection{RQ1: Does Work2Skill Improve Downstream GUI Execution?}
\label{sec:main-results}

To evaluate the effectiveness of induced skills in downstream GUI execution,
we compare \ours with No Skill and the text-adapted ShowUI-Aloha and D2T
baselines on the demonstrated tasks under shared execution settings
(Table~\ref{tab:main-results}).
Relative to the higher-scoring of the two demonstration-based baselines for
each execution model, \ours improves Score by \(3.95\)--\(7.79\) points on
MyPCBench and \(1.65\)--\(6.96\) points on the evaluated OSWorld-Verified
subset. For GPT-5.5 on MyPCBench, SR reaches \(48.44\%\), compared with
\(37.50\%\) for both Aloha and D2T.
Compared with No Skill, \ours improves the
average score by \(4.67\)--\(10.54\) points and the success rate by
\(6.25\)--\(25.00\) percentage points on MyPCBench. On the evaluated
OSWorld-Verified subset, the corresponding gains are
\(3.11\)--\(7.50\) score points and \(1.40\)--\(7.69\) percentage
points.

Efficiency gains depend on the model: \ours uses fewer execution steps
than No Skill for Opus-4.8 and both Qwen models, but more for GPT-5.5.
Skill-induction token usage is also not consistently lower than that of
the demonstration-based baselines.
Thus, \ours achieves the highest Score and SR among the compared methods
across both task sets and all four executors, with model-dependent execution
and induction costs.

\subsection{RQ2: Do Skills Generalize to New Task Instances?}
\label{sec:task-generalization}

To assess whether induced skills generalize beyond the demonstrated instance,
we evaluate the original skills on tasks with changed requirements within the
same application, without further induction.
MyPCBench Jitter contains one paired variant of each of the 64 original
tasks, changing entities, parameters, or constraints within the same
application context. For example, a flight-booking variant changes the
destination, travel date, and seat preference. GPT-5.6-sol assists in
revising task instructions and evaluation criteria, followed by human
verification (Appendix~\ref{app:jitter-settings}). Each of the four models
directly reuses the skills induced for the original tasks on their
paired variants.

\begin{table*}[ht]
\centering
\caption{
Performance on 64 MyPCBench Jitter tasks.
Bold indicates the best Score or SR within each execution model,
and blue shading denotes our method.
Smaller annotations report changes relative to No Skill, computed from
the displayed two-decimal values.
For Score and SR, upward arrows indicate improvements;
for Steps, downward arrows indicate fewer execution steps.
}
\label{tab:jitter-results}

\small
\setlength{\tabcolsep}{1.4pt}
\renewcommand{\arraystretch}{1.18}

\newlength{\mainvaluewidth}
\newlength{\deltavaluewidth}
\setlength{\mainvaluewidth}{2.5em}
\setlength{\deltavaluewidth}{2.2em}

\newcommand{\metrichead}[1]{%
    \makebox[\mainvaluewidth][c]{#1}%
    \makebox[\deltavaluewidth][l]{}%
}

\newcommand{\baseresult}[1]{%
    \makebox[\mainvaluewidth][c]{#1}%
    \makebox[\deltavaluewidth][l]{}%
}

\newcommand{\upresult}[2]{%
    \makebox[\mainvaluewidth][c]{#1}%
    \makebox[\deltavaluewidth][l]{%
        \raisebox{-0.45ex}{%
            \scriptsize
            \textcolor{teal!70!black}{$\uparrow$#2}%
        }%
    }%
}

\newcommand{\downresult}[2]{%
    \makebox[\mainvaluewidth][c]{#1}%
    \makebox[\deltavaluewidth][l]{%
        \raisebox{-0.45ex}{%
            \scriptsize
            \textcolor{red!65!black}{$\downarrow$#2}%
        }%
    }%
}

\newcommand{\stepdown}[2]{%
    \makebox[\mainvaluewidth][c]{#1}%
    \makebox[\deltavaluewidth][l]{%
        \raisebox{-0.45ex}{%
            \scriptsize
            \textcolor{teal!70!black}{$\downarrow$#2}%
        }%
    }%
}

\newcommand{\sameresult}[1]{%
    \makebox[\mainvaluewidth][c]{#1}%
    \makebox[\deltavaluewidth][l]{%
        \raisebox{-0.45ex}{%
            \scriptsize
            \textcolor{gray!70}{$\rightarrow$0.00}%
        }%
    }%
}

\begin{adjustbox}{max width=\textwidth,center}
\begin{tabular}{@{}l *{4}{ccc}@{}}
\toprule

\multirow{2}{*}{Method}
& \multicolumn{3}{c}{GPT-5.5}
& \multicolumn{3}{c}{Opus-4.8}
& \multicolumn{3}{c}{Qwen3.5-35B-A3B}
& \multicolumn{3}{c}{Qwen3.5-9B} \\

\cmidrule(lr){2-4}
\cmidrule(lr){5-7}
\cmidrule(lr){8-10}
\cmidrule(lr){11-13}

& \metrichead{Score}
& \metrichead{SR (\%)}
& \metrichead{Steps}

& \metrichead{Score}
& \metrichead{SR (\%)}
& \metrichead{Steps}

& \metrichead{Score}
& \metrichead{SR (\%)}
& \metrichead{Steps}

& \metrichead{Score}
& \metrichead{SR (\%)}
& \metrichead{Steps} \\

\midrule

No Skill
& \baseresult{73.45}
& \baseresult{37.50}
& \baseresult{52.41}

& \baseresult{83.30}
& \baseresult{45.31}
& \baseresult{51.97}

& \baseresult{42.94}
& \baseresult{12.50}
& \baseresult{67.66}

& \baseresult{27.80}
& \baseresult{6.25}
& \baseresult{82.33} \\

Aloha
& \upresult{75.50}{2.05}
& \upresult{42.19}{4.69}
& \stepdown{51.45}{0.96}

& \downresult{82.66}{0.64}
& \downresult{43.75}{1.56}
& \stepdown{48.17}{3.80}

& \downresult{41.19}{1.75}
& \sameresult{12.50}
& \stepdown{63.14}{4.52}

& \upresult{31.97}{4.17}
& \upresult{10.94}{4.69}
& \stepdown{78.63}{3.70} \\

D2T
& \downresult{73.03}{0.42}
& \downresult{35.94}{1.56}
& \stepdown{51.94}{0.47}

& \downresult{81.44}{1.86}
& \sameresult{45.31}
& \stepdown{48.48}{3.49}

& \upresult{46.92}{3.98}
& \upresult{14.06}{1.56}
& \stepdown{66.31}{1.35}

& \upresult{30.66}{2.86}
& \upresult{\textbf{12.50}}{6.25}
& \stepdown{81.66}{0.67} \\

\rowcolor{blue!8}
\textbf{Ours}
& \upresult{\textbf{81.75}}{8.30}
& \upresult{\textbf{46.88}}{9.38}
& \stepdown{49.78}{2.63}

& \upresult{\textbf{87.23}}{3.93}
& \upresult{\textbf{54.69}}{9.38}
& \stepdown{46.53}{5.44}

& \upresult{\textbf{51.58}}{8.64}
& \upresult{\textbf{21.88}}{9.38}
& \stepdown{65.17}{2.49}

& \upresult{\textbf{36.25}}{8.45}
& \upresult{\textbf{12.50}}{6.25}
& \stepdown{79.53}{2.80} \\

\bottomrule
\end{tabular}
\end{adjustbox}

\end{table*}

In Table~\ref{tab:jitter-results}, \ours improves the average score over
No Skill by \(3.93\)--\(8.64\) points
and the success rate by \(6.25\)--\(9.38\) percentage points.
Compared with the stronger demonstration-based baseline for each executor,
the Score advantage is \(4.28\)--\(6.25\) points. \ours also achieves the
highest success rate for three executors, while tying D2T for
Qwen3.5-9B. Average execution steps also decrease by
\(2.49\)--\(5.44\) relative to No Skill across the four execution models.
These gains across all four executors support reuse beyond the demonstrated
instance: the induced
skills remain useful when entities, parameters, or constraints change within
the tested applications.


\Needspace{18\baselineskip}
\subsection{RQ3: Do Skills Transfer Across Models?}
\label{sec:cross-model}

\begingroup
\setlength{\intextsep}{0pt}
\setlength{\columnsep}{0.03\textwidth}
\makeatletter\@afterindenttrue\makeatother
\begin{wrapfigure}[11]{r}{0.44\textwidth}
\centering

\includegraphics[width=\linewidth]{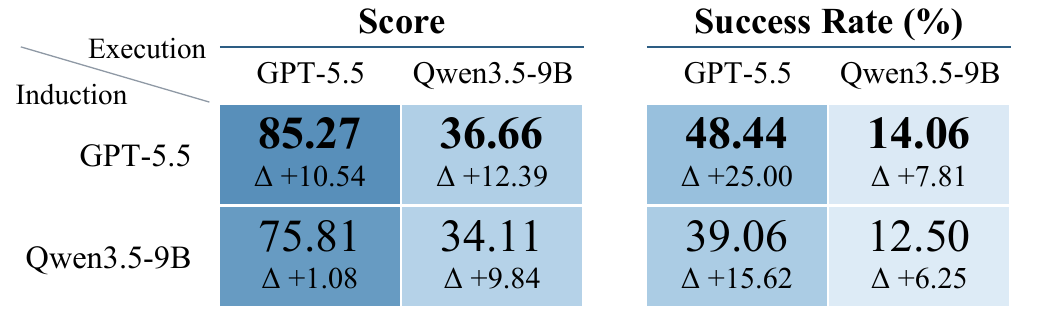}

\vspace{-5pt}

\caption{\footnotesize
Cross-model skill transfer on 64 MyPCBench tasks.
Rows denote induction models; columns denote executors.
$\Delta$ denotes the gain over the same executor's No Skill baseline,
in Score points or SR percentage points, calculated from displayed
two-decimal values.}
\label{fig:cross-model-heatmap}

\end{wrapfigure}

\noindent
To evaluate whether induced skills transfer across models, we separate
induction from execution and test guidance produced by one model with a
different executor.
We cross GPT-5.5 and Qwen3.5-9B as induction and execution models,
freezing each induced skill before downstream evaluation
(Figure~\ref{fig:cross-model-heatmap}).

Skills induced by GPT-5.5 outperform those induced by Qwen3.5-9B under
both executors. The advantage is larger with GPT-5.5 execution, reaching
\(9.46\) score points and \(9.38\) percentage points in success rate,
but remains positive after transfer to \mbox{Qwen3.5-9B}. All four pairings
improve on the No Skill baseline for the same executor.
Performance remains substantially lower with Qwen3.5-9B execution
regardless of the induction model. Thus, the skills transfer usefully between
these two models, but stronger induction does not close the
performance gap between executors on these tasks.
\par
\endgroup

\begin{figure*}[htbp]
\centering
\begin{subfigure}[t]{0.56\textwidth}
\centering
\includegraphics[width=\linewidth]{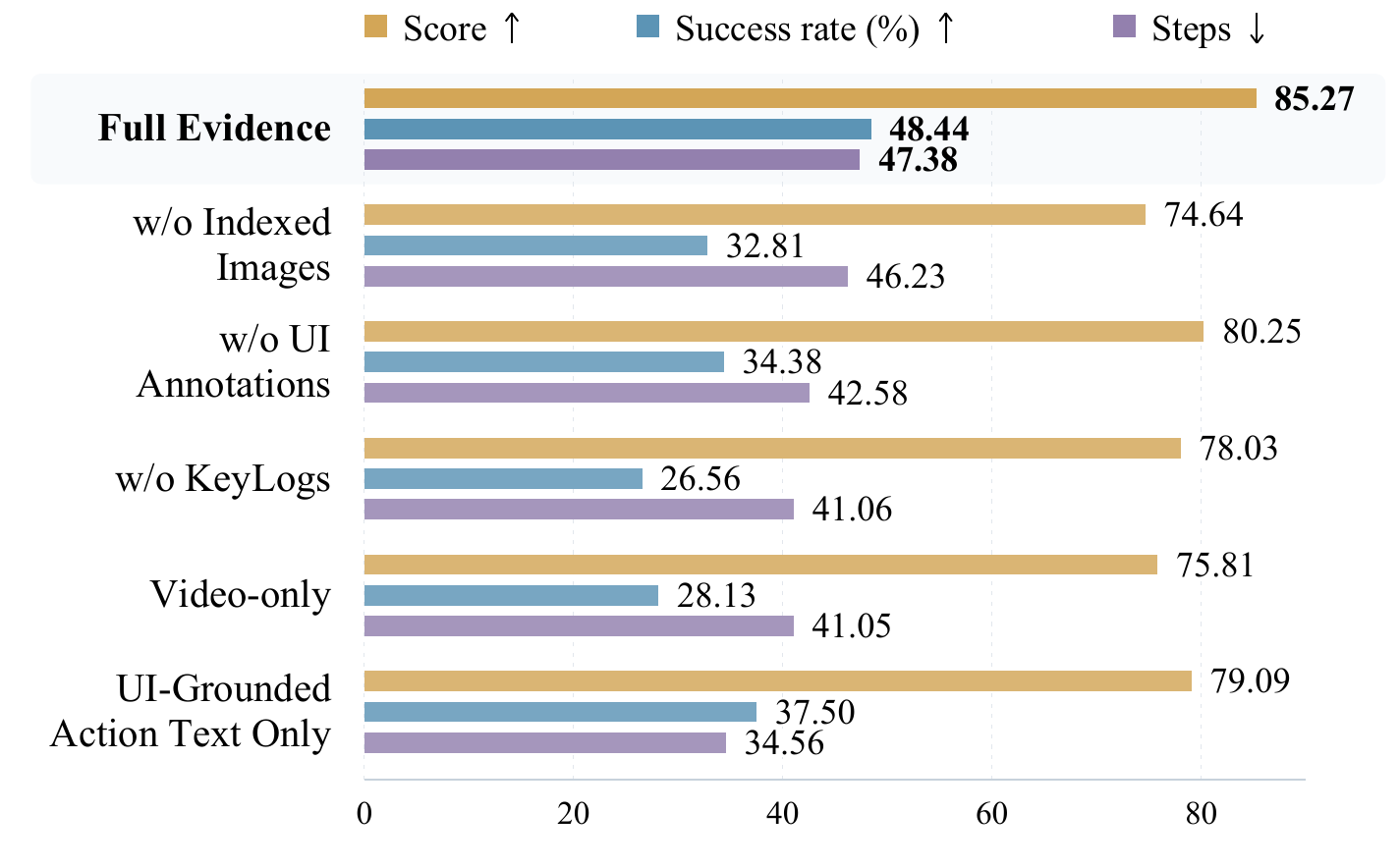}

\caption{Ablation of evidence used for skill induction.}
\label{fig:evidence-ablation}
\end{subfigure}
\hfill
\begin{subfigure}[t]{0.42\textwidth}
\centering
\includegraphics[width=\linewidth]{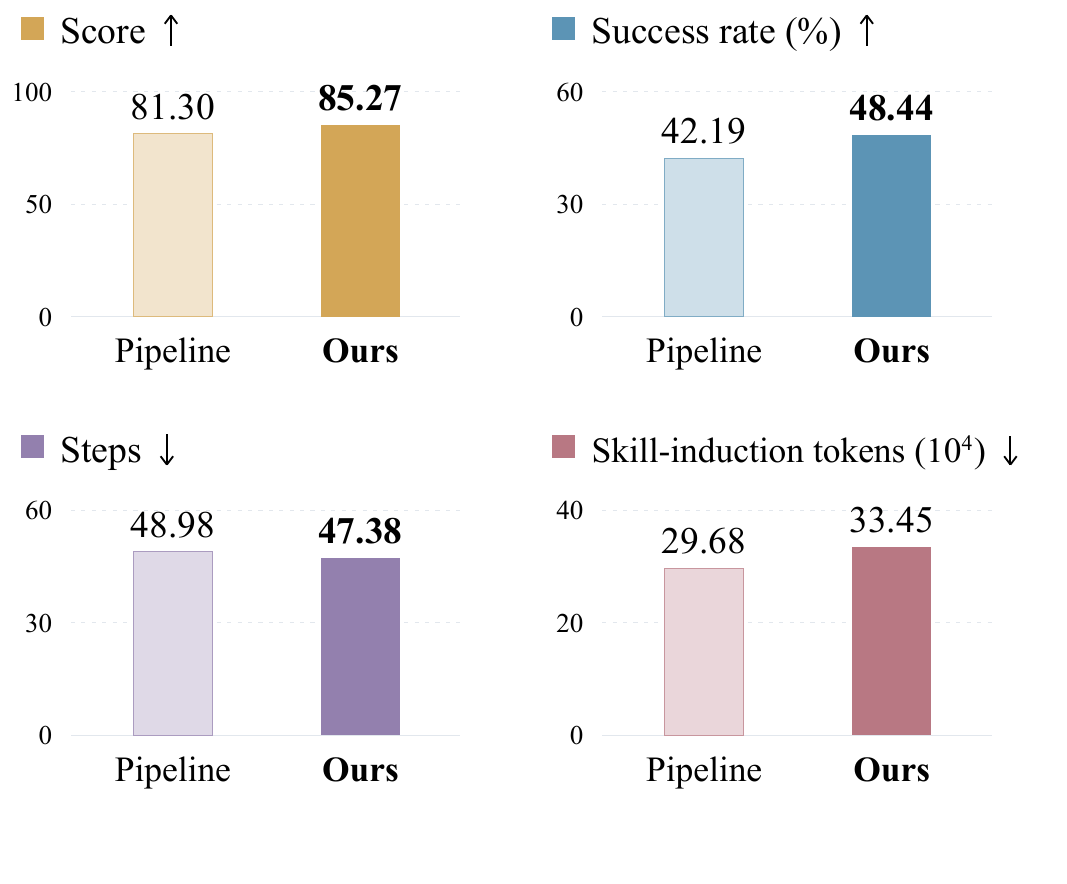}

\caption{Comparison with the pipeline prompt.}
\label{fig:pipeline-comparison}
\end{subfigure}

\vspace{-2pt}

\caption{
Ablation and pipeline-prompt comparison on 64 MyPCBench tasks using
GPT-5.5 with high reasoning effort. In (a), Full Evidence uses all available
modalities; UI-Grounded Action Text Only combines recorded actions with target UI
descriptions. w/o KeyLogs also removes overlays from indexed images.
Score, SR, and Steps measure downstream execution; Tokens
reports mean skill-induction token usage per task in units of \(10^4\).
}
\label{fig:ablation-and-prompt}

\vspace{-10pt}
\end{figure*}

\subsection{RQ4: How Do Evidence and Induction Prompts Affect Skill Induction?}
\label{sec:induction-ablation}

To assess how evidence and prompts affect downstream skill utility,
we ablate the available input modalities and separately compare the \ours
and Pipeline Prompt system prompts.
For the evidence ablation, we vary induction evidence on 64 MyPCBench tasks
with GPT-5.5, keeping
reasoning effort (high) and the iteration limit fixed. UI annotations contain
metadata and descriptions derived from the corresponding action targets in
the AX tree. The w/o KeyLogs condition removes both raw logs and added image
overlays while retaining the action trace and its UI target information.
Appendix~\ref{app:modality-settings} details the inputs and prompt adaptations.
Full Evidence scores \(85.27\) and fully solves 31 tasks
(Figure~\ref{fig:evidence-ablation}). Removing indexed images, UI annotations,
or KeyLogs reduces Score by
\(5.02\)--\(10.63\) points and SR by \(14.06\)--\(21.88\) percentage points.
The largest drops are in Score without indexed images and SR without KeyLogs.
UI-Grounded Action Text Only uses ordered action descriptions with target UI
semantics. It outperforms Video-only and uses the fewest steps;
both trail Full Evidence in Score and SR.
Fewer steps do not necessarily indicate better task completion;
Score and SR remain essential for interpreting execution efficiency.










With the evidence held fixed, we next test whether a prescribed sequence of
induction stages helps the agent use it. Our concise default prompt leaves
detailed evidence exploration and skill revision to the agent. Pipeline Prompt
specifies the induction stages and their order: read the task and artifacts,
inspect images and video,
then write, review, and finalize skills. Only the system prompt changes;
the model, evidence, tools, authoring policy, and output format remain fixed
(Appendix~\ref{app:pipeline-prompt}). Both prompts allow further evidence
inspection. In Figure~\ref{fig:pipeline-comparison},
\ours improves Score by \(3.97\) points and SR by \(6.25\) percentage points,
while reducing average execution steps from \(48.98\) to \(47.38\), at
\(12.7\%\) higher skill-induction token usage.
In this setting, Full Evidence yields the best Score and SR among the tested
input configurations, while the \ours prompt outperforms the prescribed
pipeline at higher induction cost. Together, these comparisons show how
evidence and prompt design affect the downstream utility of generated skills.






\vspace{-10pt}
\begin{figure}[htbp]
    \centering
    \includegraphics[width=\linewidth]{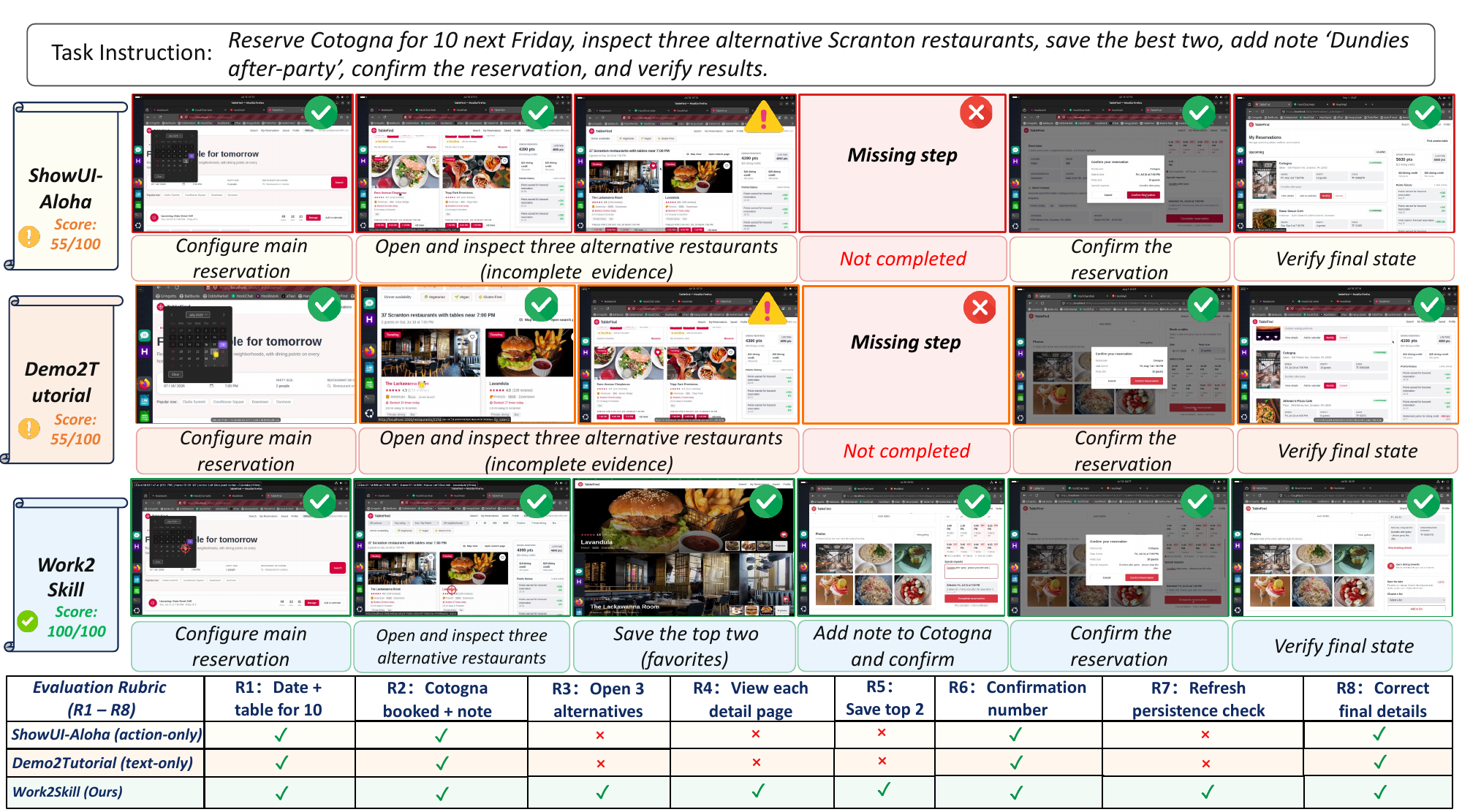}
    \caption{
    Execution trajectories (top) and rubric outcomes (bottom) for a TableFind
    reservation with fallbacks, using GPT-5.5 with each method's extracted
    guidance. Both baselines leave fallback handling and persistence checks
    incomplete (\(55/100\)); \ours satisfies all eight requirements
    (\(100/100\)). Appendix~\ref{app:case-analysis} provides the detailed rubric
    and execution analysis.
    }
    \label{fig:case-study}
    \vspace{-10pt}
\end{figure}

\subsection{Case Study}
\label{sec:case-study}

To illustrate how extracted skill guidance affects execution, we compare
GPT-5.5 supplied with outputs from ShowUI-Aloha, D2T, or \ours through
the shared text interface (Section~\ref{sec:experimental-setup}). Each method
extracts its guidance from the same human demonstration.
Figure~\ref{fig:case-study} shows a Cotogna booking for ten people with
three fallback restaurants and two favorites. The fallbacks must be handled
before confirmation, while preserving the date, party size, and requested note.
With ShowUI-Aloha or D2T guidance, the executor completes the primary booking
but leaves fallback inspection, favoriting, and persistence checks incomplete,
scoring \(55/100\) in both conditions. With the skill induced by \ours, it
satisfies all eight rubric items (\(100/100\)).
The generated skill makes intermediate requirements explicit: it rechecks
fields that may reset during navigation, excludes the
primary restaurant from the fallback set, and defers final confirmation until
the fallback steps are complete. The skill preserves dependent subgoals across
navigation and checks both the reservation and saved favorites before declaring
completion
(Appendix~\ref{app:case-analysis}).

\section{Conclusion}
\label{sec:conclusion}
We presented \ours, a framework for capturing human GUI experience and
converting it into reusable agent skills. WorkRecorder collects aligned
multimodal evidence, WorkDemo-350 organizes 350 curated demonstrations, and
agentic skill induction combines on-demand evidence inspection with iterative
skill writing and revision. Across four execution models, \ours achieves the
highest scores and success rates among the compared methods on MyPCBench
Bounded Action and the evaluated OSWorld-Verified subset. Further experiments
support reuse under changed task requirements within the same application and
transfer between the tested models. These results highlight the value of
aligned human demonstrations and flexible evidence exploration for building
reusable procedural guidance that computer-use agents can apply through a shared
text interface without accessing the original recordings or other demonstration
evidence.

\bibliography{iclr2027_conference}
\bibliographystyle{iclr2027_conference}

\clearpage
\appendix
\section{Recorder Implementation}
\label{app:recorder}

\subsection{Architecture and Deployment}
WorkRecorder consists of a native macOS recording application, a local service
for capture coordination and storage, and accessibility providers. In local
macOS mode, the application records the selected screen region and input
activity, with available accessibility evidence supplied by the macOS provider.
For a remote Linux workflow, recording remains on the Mac, while an AT-SPI
collector runs inside the target desktop session. The Linux launcher connects
this collector to the local service through an SSH tunnel; the remote-desktop
client supplies the displayed workflow. Task initialization and official
benchmark evaluation are handled separately from recording.

\paragraph{Video and input capture.}
The macOS application records a selected screen region through ScreenCaptureKit
and encodes video with H.264. CoreGraphics event taps supply mouse and keyboard
activity, which is also grouped into semantic operations. The operator can
pause, resume, stop, and review a recording. Full keystroke capture is disabled
by default; when enabled, it records key presses and releases, characters,
modifiers, and timestamps. Otherwise, ordinary typing is represented through
aggregate typing events. The capture policy therefore determines the detail
available in the input stream.

\paragraph{Accessibility evidence.}
The Linux collector supports X11/Xorg desktops and uses AT-SPI to obtain
Accessibility Trees, interface events, and target UI elements. Its default
periodic tree sampler runs every two seconds, with limits of six levels and
400 nodes; these are implementation defaults rather than dataset collection
settings. Events may reference the latest completed tree snapshot, with its
sampling timestamps. Target-element queries return available names, roles,
and bounds at a screen location. The recorder maps positions in the displayed
remote-desktop region to Linux desktop coordinates for these queries.
Availability depends on application accessibility support and sampling limits.

\paragraph{Target-element association.}
When resolving clicks from accessibility snapshots, the resolver considers nodes
whose bounds contain the click coordinate. It ranks these candidates by node
confidence and interactive role, with an area penalty favoring smaller controls.
Resolution records retain the selected source, confidence, and evidence anchor;
uncertain cases can remain flagged for review.

The remote Linux probing path is triggered by clicks and drags. Configured
remote-display geometry maps local pointer positions to Linux desktop pixels;
positions outside the displayed desktop do not trigger a point probe. The
AT-SPI watcher combines bounded point queries with nearby focus, selection, and
state-change events. Its point-query ranking prioritizes active windows, named
interactive controls, other named nodes, greater depth, and smaller area, in
that order. Event correlation normally checks whether the target bounds contain
the pointer position, with additional handling for transient controls such as
menus that disappear after selection. Pointer-resolution records retain the
originating pointer sequence identifier and distinguish resolved, coarse,
unresolved, and error or timeout outcomes.

During export, recorded probe results are associated with actions using their
trigger's monotonic timestamp and action kind, with a nearest-action fallback
within 2.5 seconds. In normal export mode, a missing click or drag target can
also be recovered from the nearest specific accessibility event within a
3-second window. This backfill excludes generic page or window focus and uses
each candidate event at most once. For drags, the matching anchor is the drag
start, computed from the release time and recorded duration. Backfilled records
retain their source, event identifier, and time offset. These are heuristic
temporal associations; the matching windows are implementation tolerances, not
measured accuracy bounds. Keyboard input does not trigger this Linux target
probe. AX snapshots are attached independently as interface context.

\subsection{Keyframe Selection}
\label{app:keyframes}
\paragraph{Capture triggers.}
WorkRecorder selects keyframes primarily around recorded operations and
interface transitions. Recording start, stop, pause, and resume provide boundary
frames, while users can explicitly mark additional frames. Routine candidates
are triggered by clicks, drag releases, scroll completion, recognized hotkeys,
command submission, and window or tab switches. A typing pause can trigger a
frame when text is submitted or a typing milestone is reached; the default
milestone requires at least eight characters or input events in the typing
segment. Individual key presses and releases do not independently trigger
screenshots. Additional enabled policies can request frames at scroll
milestones or when a provider reports an error or warning surface.

\paragraph{Action context and capture timing.}
For clicks, hotkeys, and window or tab switches, the recorder can reuse a recent
preceding keyframe as before-action context, with a default freshness limit of
1.5 seconds. A new before-action capture requires an explicit pre-action event;
otherwise this context can remain unavailable. Result frames use a bounded
settling window to allow the interface to respond. With adaptive polling
enabled, capture can proceed when a provider reports readiness or when visual
or text observations stabilize after a detected change; the default stability
rule requires two consecutive stable samples. Waiting limits depend on the
action type. If settling times out, the policy can capture the available state
and flag it for review, or suppress the candidate. Thus, a result-frame link
does not by itself establish that the interface has finished updating.

\paragraph{Candidate filtering.}
The planner removes duplicate triggers within an action's timing plan. The
scheduler applies a minimum gap between separate plans, limits result frames
per action, and enforces capture budgets. When change or quality signals are
available, it can also reject candidates with insufficient change or
information gain. Defaults include a 500-ms gap, at most two result-related
candidates per action, and a session budget of 64 keyframes; required boundary
frames and explicit user marks have exemptions. These are implementation
defaults, not verified settings for every collected demonstration. Retained
keyframes carry identifiers and timestamps, while trigger decisions retain
acceptance or rejection reasons. Frame reuse and candidate filtering mean that
actions and keyframes need not have a one-to-one correspondence. Exported image
counts are distinguished from recorded keyframe counts in
Appendix~\ref{app:schema}.

\subsection{Time Alignment}
The recorder maps evidence onto the stitched video's timeline. Native input
uses a monotonic timestamp anchored to the first captured frame of each
recording segment, rather than the time at which recording was requested.
Remote Linux events use wall-clock timestamps, which are projected onto the
same timeline using the corresponding first-frame wall-clock anchors and
segment durations. Segment offsets account for pause/resume boundaries; remote
events falling within a pause have no corresponding video frame. The Linux
launcher checks host clock agreement before recording and rejects a difference
greater than two seconds. This check is a deployment condition, not a measured
alignment-error bound. Legacy recordings without first-frame wall-clock anchors
use the session start as a fallback.

\section{Evidence Schema}
\label{app:schema}

\subsection{Episode Files for Skill Induction}
The induction agent reads a per-task episode directory containing the task and
its recorded evidence. The benchmark converter creates three core files:
\texttt{task.json}, \texttt{episode.json}, and \texttt{trace/actions.jsonl}.
Input logs, video, summaries, and images are included when available
(Table~\ref{tab:task-schema}). The task file must contain the authoritative
instruction and task inputs; a placeholder with only an inferred task ID is
insufficient for induction. The converter excludes the \texttt{grading} field
from this file.

The \texttt{episode.json} manifest identifies the task and recording, records
duration and evidence counts, and indexes the available files. Evidence paths
are relative to the episode directory. Its \texttt{source} entry records the
original export location for provenance; that location is not an additional
input root for the agent.

\begin{table}[H]
    \centering
    \caption{Files exposed to the induction agent in a converted episode.
    The first three files form the core structure; the remaining artifacts
    are included when available.}
    \label{tab:task-schema}
    \small
    \begin{tabularx}{\linewidth}{@{}p{0.39\linewidth}X@{}}
        \toprule
        File or directory & Content \\
        \midrule
        \texttt{task.json} & Task instruction and associated task metadata. \\
        \texttt{episode.json} & File references, recording metadata, and counts. \\
        \texttt{trace/actions.jsonl} & Normalized action records in temporal order. \\
        \midrule
        \texttt{trace/keylog.jsonl} & Recorded input-event log. \\
        \texttt{video.mp4} & Recording video. \\
        \texttt{summary.md} & Textual operation summary. \\
        \texttt{evidence/} & Evidence images, with available image manifests
        and action-association indices. \\
        \bottomrule
    \end{tabularx}
\end{table}

\subsection{Action Records and Image Associations}
Each line of \texttt{trace/actions.jsonl} describes a recorded action, with
its identifier, sequence index, action kind, video-relative timestamp, and
caption. Coordinates and target information are retained when available.
The following excerpt reproduces selected fields from the first action in
the \texttt{hard\_app-f016} episode used in Appendix~\ref{app:examples}:
\begin{small}
\begin{verbatim}
{
  "sequence_index": 0,
  "action_id": "act_0JYEZK480Ta_81AR",
  "kind": "click",
  "media_time_ms": 3908,
  "caption": "Left Click push button - Calendar",
  "target_label": "push button - Calendar"
}
\end{verbatim}
\end{small}
When present, \texttt{evidence/manifest.json} and
\texttt{evidence/image\_actions.jsonl} provide the image index and its input-event
and trajectory associations. In this episode, the first image-index record
references the same \texttt{action\_id}, with \texttt{frame\_time\_ms = 3908}
and \texttt{action\_delta\_ms = 0}. These fields let the agent locate the
corresponding event frame and nearby video observations. A nonzero
\texttt{action\_delta\_ms} indicates an approximate temporal match. An event
frame captures the interface near the input event; verifying its effect may
require a subsequent observation.

\subsection{Evidence Availability and Counts}
The schema describes the files available within an episode, rather than
guaranteeing every modality for every demonstration. Target metadata in an
action record does not imply that a complete accessibility-tree sequence is
included. In Linux mode, the continuous tree sequence remains in the Linux
evidence directory, while the Mac export can preserve snapshots received with
events. The induction agent can inspect only the evidence provided within its
exposed input directory.

The manifest distinguishes action records, input events, recorded keyframes,
and exported evidence images. Table~\ref{tab:data-stats} reports mean counts of
action records and exported images per episode. The \texttt{hard\_app-f016}
episode, for example, contains 37 action records and 23 evidence images,
alongside 176 input events and 29 recorded keyframes.

Action and image counts differ because they use separate grouping rules.
Image export pairs key presses with releases, groups double clicks, and merges
adjacent typing evidence into one image with the last frame and a combined text
annotation. Scroll records do not independently generate images, while action
preprocessing merges adjacent typing and scroll records.

The converter sets \texttt{counts.keyframes} from the recording's keyframe
count, falling back to the number of loaded keyframe records;
\texttt{counts.evidence\_images} counts images copied into the main evidence
collection. Exporting all available keyframe screenshots is a separate option
that creates \texttt{evidence/all\_keyframes/} and its own manifest. Recorded
keyframes thus reflect the capture policies in Appendix~\ref{app:keyframes};
exported images reflect subsequent evidence processing.

\Needspace{24\baselineskip}
\section{Evidence-Exploration Tool Definitions}
\label{app:tools}

\begin{table}[H]
    \centering
    \caption{Implemented induction tools. NexAU additionally provides
    \toolname{LoadSkill} for loading the bundled authoring policy.}
    \label{tab:evidence-tools}
    \begin{tabular}{p{0.32\linewidth}p{0.60\linewidth}}
        \toprule
        Tool & Capability \\
        \midrule
        \toolname{list\_directory} & List files and subdirectories within an
        exposed directory. \\
        \toolname{glob} & Find files matching a glob pattern within the
        exposed input or output directory. \\
        \toolname{read\_file} & Read text and structured evidence with line
        numbers and pagination. \\
        \toolname{read\_visual\_evidence} & Return an ordered batch of
        1--20 images labeled with their source paths. \\
        \toolname{get\_video\_metadata} & Inspect video duration, dimensions,
        frame rate, codec, and file size. \\
        \toolname{watch\_video} & Sample timestamped frames over a selected
        interval, subject to a frame limit. \\
        \toolname{extract\_video\_frames} & Request frames at selected
        millisecond timestamps. \\
        \toolname{write\_file} & Create or replace a complete output file. \\
        \toolname{finalize\_extraction} & Check output structure and tool-use
        requirements, then write the extraction manifest on success. \\
        \bottomrule
    \end{tabular}
\end{table}

The input directory is exposed as \texttt{.} and generated files as
\texttt{output/}. File tools resolve paths within these roots; writes are limited
to the output directory and cannot replace runtime-owned manifests or logs.
Task interpretation, action selection, and comparisons between observations
are performed by the agent using the returned evidence. The following
descriptions summarize the registered tools and their principal arguments.

\paragraph{File discovery and reading.}
\toolname{list\_directory} takes \texttt{dir\_path}; \toolname{glob} takes a
\texttt{pattern} and optional \texttt{dir\_path}. Both support filtering options.
\toolname{read\_file} takes \texttt{file\_path}, with optional zero-based line
\texttt{offset} and \texttt{limit}. It handles text, JSON, JSONL, YAML, and
Markdown as text, returns numbered lines and continuation guidance, and directs
image and video requests to the corresponding visual tools. A response is
limited to 20,000 characters of numbered content; individual long lines are
truncated. This interface provides access to task metadata, action traces,
input logs, image indices, and any exported structured UI artifacts.

\paragraph{Image inspection.}
\toolname{read\_visual\_evidence} takes an ordered \texttt{file\_paths} list
containing 1--20 distinct images. Optional \texttt{image\_detail} and
\texttt{image\_max\_size} control detail and resizing. Returned images preserve
the requested order and have source-path labels, allowing the agent to compare
selected states while retaining their file identities.

\paragraph{Video inspection.}
\toolname{get\_video\_metadata} takes \texttt{file\_path} and reports metadata
obtained with \texttt{ffprobe}. \toolname{watch\_video} takes the video path and
optional \texttt{start\_ms}, \texttt{end\_ms}, \texttt{interval\_ms},
\texttt{max\_frames}, and \texttt{frame\_width}. It returns sampled images;
defaults are a 5,000-ms interval, at most 12 frames, and a width of 1,280 pixels.
The interval must be at least 100 ms and each call can request at most 20 frames.
\toolname{extract\_video\_frames} instead takes a list of 1--20
\texttt{timestamps\_ms} and an optional \texttt{frame\_width}. Both use
\texttt{ffmpeg} and return timestamp labels. Near the video end, the requested
time can be adjusted backward to obtain a decodable frame; the label reports
the sampling time used and the original request when adjusted. Finalization
checks that metadata inspection preceded sampling for the same video path.

\paragraph{Writing and finalization.}
\toolname{write\_file} takes \texttt{file\_path} and the complete \texttt{content},
creating or replacing a file. \toolname{finalize\_extraction} takes no required
arguments. It requires at least one skill, a recorded read of an input
\texttt{task.json}, valid skill frontmatter and a nonempty body, a
\toolname{write\_file} call and a subsequent \toolname{read\_file} call for every
final skill, and the video-metadata ordering described above. The skill name
must use lowercase hyphen-case and match its folder; its description must be
nonempty. Heuristic checks for fixed coordinates, local paths, and timestamps
produce warnings rather than semantic correctness guarantees.

\Needspace{7\baselineskip}
\subsection{Example Exploration Trace}
An illustrative sequence using these tools is:
\begin{enumerate}
    \raggedright
    \item load \texttt{skill-creator}, locate the authoritative
    \texttt{task.json}, and read it with \toolname{read\_file};
    \item discover evidence files with \toolname{list\_directory} and
    \toolname{glob}, then read relevant traces and image indices;
    \item inspect representative initial, critical, and final-state images
    with \toolname{read\_visual\_evidence};
    \item if a transition remains unclear, read the video's metadata and use
    \toolname{watch\_video} or \toolname{extract\_video\_frames} to examine it;
    \item write a skill with \toolname{write\_file}, inspect further evidence
    if needed, and revise the draft; and
    \item read the final \texttt{SKILL.md} after its last write and call
    \toolname{finalize\_extraction}.
\end{enumerate}
This is an example, not a prescribed sequence of all tool calls. The system
prompt sets the task-first and representative-image requirements; the agent
chooses the detailed exploration and revision sequence. On successful
finalization, the runtime manifest records skill hashes, inspected source paths,
validation results, and provenance. Separate run logs capture tool activity and
available usage information. These records provide run-level provenance without
asserting a source citation for every generated skill statement.

\section{Prompts}
\label{app:prompts}

\subsection{Induction System Prompt}
The induction system prompt sets the task, evidence-access boundaries, and
completion requirements. Its instructions are reproduced below.
\begin{promptbox}{Induction System Prompt}
You extract reusable NexAU skills from one recorded computer-use benchmark episode.

\begin{promptitems}
\item Stay within the exposed input and output roots. Never seek grading
rubrics, judge output, reference answers, or files outside them.
\item First load \texttt{skill-creator}; it is the sole authority for Skill
boundaries and content. Do not restate, shorten, or weaken its authoring rules.
\item Read the authoritative \texttt{task.json} before interpreting the
trajectory. Treat traces and recordings as fallible evidence; recording
completion is not task success.
\item When indexed evidence images exist, inspect representative initial,
critical, and final-state frames. Before sampling any video, call
\toolname{get\_video\_metadata} for that exact path.
\item Write generated Skill folders only under the output root. Do not create
or edit runtime-owned \texttt{extraction-manifest.json},
\texttt{run-events.jsonl}, \texttt{run-metadata.json}, or
\texttt{run-summary.json}.
\item Read each final \texttt{SKILL.md} after its last write, then call
\toolname{finalize\_extraction}. Correct every reported error and retry. Only
successful finalization completes the task.
\end{promptitems}
\end{promptbox}

\subsection{NexAU Agent Configuration and Completion}
The launcher creates one NexAU agent with the system prompt, registered tools,
and bundled \texttt{skill-creator} policy. Its initial message directs the agent
to load that policy, read the task, create skills under the output root, read
back each final skill, and finalize. The implementation uses structured tool
calls; it does not prescribe a separate controller state or a numerical
information-gain objective. Evidence inspection and skill authoring occur in
the same agent conversation.

The model and endpoint are supplied at launch, with configurable iteration,
context, output, temperature, and reasoning settings. Our experiment settings
are reported in Appendix~\ref{app:model-settings}. Provider-request
retries in the supplied implementation are configured with 30 attempts, a
maximum backoff of 30 seconds, and a 900-second request timeout.

The prompt instructs the agent to correct finalization errors and retry.
\toolname{finalize\_extraction} returns diagnostics on failure and atomically
writes \texttt{extraction-manifest.json} on success. The launcher checks that
finalization succeeded after the agent returns; an ordinary text response alone
leaves the run incomplete. The runtime also writes configuration provenance,
event logs, and a summary of status, timing, tool activity, and available token
usage. Structural validation and recorded call ordering are enforced by code;
the authoring policy's requirements for evidence grounding, workflow completeness,
and appropriate generalization remain instructions to the model.

\subsection{Pipeline Prompt}
\label{app:pipeline-prompt}
The Pipeline Prompt baseline changes only the induction system prompt, keeping
the model, input evidence, tools, skill-authoring policy, and output format the
same as \ours. It expands the concise bootstrap prompt into an explicit workflow
for reading the task, inventorying artifacts, inspecting structured evidence and
images, sampling video, and authoring and validating skills. Both conditions
load the same \texttt{skill-creator} policy, read back final skill files, and
call \toolname{finalize\_extraction}.

The detailed prompt prescribes an inspection order and frame-selection rules,
including a focused initial image batch and further inspection when ambiguity
remains. One possible explanation for the prompt-ablation results
(Section~\ref{sec:induction-ablation}) is that a predefined exploration route
may narrow the range of evidence the model considers. The baseline prompt is
reproduced below, with formatting adapted for readability.

\begin{promptbox}{Pipeline Prompt}
You extract reusable NexAU skills from recorded computer-use benchmark artifacts.

\promptheading{Boundaries}
\begin{promptitems}
\item The tools expose one logical input root and one logical output root.
Treat those boundaries as absolute.
\item Never seek grading rubrics, judge output, reference answers, or files
outside the exposed roots.
\item The runtime owns \texttt{extraction-manifest.json},
\texttt{run-events.jsonl}, \texttt{run-metadata.json}, and
\texttt{run-summary.json}. Never create, edit, or overwrite those files.
\item A normal text response does not complete the task. Only a successful
\toolname{finalize\_extraction} call completes it.
\item The task instruction is authoritative. Recording status only means
recording finished; it does not prove task success.
\item Read the authoritative task instruction before interpreting the
trajectory. Preserve its goals, quantities, exclusions, ordering constraints,
and requested scope while parameterizing episode-specific values.
\item A demonstrated state change teaches interface mechanics but does not
authorize a broader goal or an additional persistent mutation.
\item Treat screenshots, action logs, key logs, summaries, and video as
fallible evidence that must be reconciled.
\item Before the first \toolname{watch\_video} or
\toolname{extract\_video\_frames} call for any video path, call
\toolname{get\_video\_metadata} for that exact path. This ordering is irreversible
within a run: calling metadata after sampling cannot repair it, and runtime
event logs must never be edited.
\end{promptitems}

\promptheading{Image Evidence}
\begin{promptitems}
\item Files under an episode\textquotesingle s \texttt{evidence/} directory are
event-aligned evidence frames, not arbitrary screenshots. They are decoded from
the episode video at \texttt{frame\_time\_ms}, which is derived from a grouped
key or pointer event after applying \texttt{frame\_offset\_ms}.
\item Read \texttt{evidence/manifest.json} or
\texttt{evidence/image\_actions.jsonl} before selecting images. These indexes
connect each image to its event kind, event time or interval, pointer points,
and the nearest trajectory \texttt{action\_id} and caption. A nonzero
\texttt{action\_delta\_ms} means the event and trajectory action were only
approximately aligned.
\item The image annotation is generated evidence metadata, not part of the
recorded application UI. Its top banner describes the input event and matched
trajectory action. Click and double-click frames mark the pointer target; drag
frames show the recorded path; typing frames are taken at the end of the grouped
typing interval and describe the recorded keys.
\item Use these full-resolution frames to identify visible labels and semantic
controls, verify which item or field was targeted, read entered or selected
values, understand drag endpoints, and compare UI state across important
actions. They are especially useful where coarse video sampling makes text or
state changes ambiguous.
\item An event frame shows the UI at or near the input event. Do not assume it
proves the action\textquotesingle s resulting state. Verify cause and effect by
comparing the next relevant evidence frame or by extracting a video frame
shortly after the event. Because images are event-triggered, the final
successful state may require inspection near the end of the video.
\item Pointer coordinates, timestamps, generated crosshairs or paths, and one
episode\textquotesingle s action caption are trace evidence only. Convert them
into semantic targets and observable state; never copy them into a reusable
skill as fixed instructions.
\item If an episode contains evidence images, inspect 1--20 representative
images together with \toolname{read\_visual\_evidence}. Cover the initial state,
critical choices or inputs, important transitions, and the final observable
outcome. \toolname{watch\_video} samples do not replace this full-resolution
inspection.
\end{promptitems}

\promptheading{Workflow}
\begin{promptsteps}
\item Locate and read the authoritative \texttt{task.json} before interpreting
the trajectory, then inventory the input recursively with
\toolname{list\_directory} and \toolname{glob}. Pass \texttt{dir\_path: "."} for
the logical input root; use \texttt{dir\_path: "output"} for generated files.
\item Read textual and structured artifacts with \toolname{read\_file}. The same
tool handles text, JSON, JSONL, YAML, and Markdown.
\item Read the evidence index, select representative initial, critical, and
final event frames, and inspect them with \toolname{read\_visual\_evidence}.
Prefer one focused batch and inspect more only when ambiguity remains. A
one-item \texttt{file\_paths} list reads a single frame; the tool accepts up to
20 paths.
\item For video, call \toolname{get\_video\_metadata} for the exact video path
before any sampling. Only then use \toolname{watch\_video} for coarse or local
interval sampling and \toolname{extract\_video\_frames} at exact action timestamps
when details matter.
\item Load \texttt{skill-creator} before authoring output.
\item Infer reusable procedures, semantic target-selection rules, observable
completion checks, and supported recovery behavior. Separate these from
episode-specific values and incidental human actions. Keep phases in one
composite Skill when they share state, task-derived data, requested counts, or
required order. Split only independently invocable capabilities.
\item Write each skill as \texttt{<skill-name>/SKILL.md} under the output root.
Add references or scripts only when they directly improve reuse.
\item Before finalizing, read every final \texttt{SKILL.md} back with
\toolname{read\_file} after its last write and self-review it:
\begin{promptitems}
\item \texttt{SKILL.md} starts with YAML frontmatter containing a lowercase
hyphen-case \texttt{name} matching its folder and a non-empty
\texttt{description}.
\item The body is non-empty, reusable, and uses semantic targets rather than
fixed coordinates.
\item Every explicit task goal, constraint, requested quantity, selection rule,
scope, and required ordering relationship needed at runtime is present in the
Skill.
\item Every persistent state change instructed by the skill is authorized by
the task.
\item It contains no machine paths, episode-specific timestamps, action
indexes, session IDs, or instance values presented as general procedure.
\item Any material uncertainty is expressed in the Skill as an observable
conditional check or limitation.
\end{promptitems}
\item Call \toolname{finalize\_extraction}. It confirms that \texttt{task.json}
was read, every final Skill was read back after its last write, video metadata
ordering was respected, and all final Skill files are valid. It writes the
runtime-owned manifest. If it reports any problem, correct the output and call
it again. Do not finish with an ordinary text response.
\end{promptsteps}

\promptheading{Quality Rules}
\begin{promptitems}
\item Do not encode fixed screen coordinates, action sequence indexes,
timestamps, local machine paths, session IDs, or one recording\textquotesingle s
dates, names, booking IDs, and other instance values as general procedure.
\item Prefer semantic UI roles, visible labels, relative state, and task-derived
selection criteria.
\item Omit actions that appear to be retries, misclicks, idle behavior, or
exploratory scrolling unless they reveal a reusable recovery strategy.
\item State only what the instruction or observable artifacts support. Express
material uncertainty as an observable check instead of inventing hidden success
criteria.
\item Use the minimum sufficient operational detail. The Skill must remain
executable without the source recording or extraction manifest.
\end{promptitems}
\end{promptbox}

\subsection{Skill-Creator}
\label{app:skill-output}
\begin{promptbox}{Skill-Creator}
\begin{promptcode}
\texttt{---}\par
name: skill-creator\par
description: Create self-contained, operationally complete skills from inspected
computer-use benchmark evidence. Use when an extractor must convert task
instructions, action traces, screenshots, and video observations into portable
skill folders without copying episode-specific details.\par
\texttt{---}\par
\end{promptcode}

\promptheading{Create Reusable Skills}
Convert observed task behavior into instructions that another agent can
execute with fresh task inputs. Describe stable decisions, semantic targets,
visible state, completion checks, and supported recovery rather than replaying
the recorded trajectory. This Skill is the authoring policy for generated
Skills; do not replace its requirements with a shorter summary.

\promptheading{Runtime Contract}
The execution agent receives the generated Skill and its bundled resources,
not the source recording, action trace, or extraction manifest. Put every
correctness-critical instruction in \texttt{SKILL.md}; supporting references must not
be required to reconstruct an omitted workflow, formula, selection rule,
quantity, ordering constraint, stop condition, or completion check.

For a graphical computer-use task, preserve the interaction modality. Instruct
the agent to make task-state changes through visible application controls and
verify them in the visible application. Never substitute an API, database edit,
script, browser console, developer tool, or other hidden state mutation unless
the authoritative task explicitly requires that mechanism.

Length is not a quality target. Remove generic explanation and trace
provenance, not operational invariants. If removing a detail could change the
selected target, computed value, action order, affected quantity, visible
success state, or recovery behavior, retain it.

\promptheading{Define the Skill Boundary}
Create one Skill for each coherent reusable capability. Keep phases together
when they exchange task-derived data, share application state or counts, or
must execute in a particular order. Split Skills only when each capability is
independently invocable and useful; do not fragment one workflow merely to make
individual files shorter.

Choose a short lowercase hyphen-case name under 64 characters. Use the same
name for the output folder and the frontmatter \texttt{name}.

\promptheading{Separate Procedure from Trace}
Treat the task instruction as the goal and the recording as fallible evidence.
Read the task first, then reconcile task data, action logs, evidence images,
and video before stating the reusable procedure. Recording completion alone
does not prove task success.

A recorded state change demonstrates interface mechanics but does not authorize
an extra persistent mutation. Preserve only state changes required by the task,
and retain:
\begin{promptitems}
\item semantic application, page, dialog, field, item, and control names;
\item rules for choosing the correct target from changing task inputs;
\item exact formulas, operands, units, basis, thresholds, rounding rules, and where
each computed result is used;
\item requested quantities and scope, including per-item repetition and count-based
stop conditions;
\item dependencies, required order, and data or state passed between phases;
\item visible checks after critical mutations and a final observable completion
check, including refresh or reopen checks when persistence matters;
\item evidence-supported recovery from loading delays, stale searches, hidden
controls, misclicks, dialogs, validation errors, and partial completion.
\end{promptitems}

Remove screen coordinates, action indexes, timestamps, session identifiers,
local machine paths, and retries or idle actions that do not teach recovery.
Replace one episode\textquotesingle s names, dates, IDs, routes, values, filenames,
and messages with task-derived parameters when they are inputs rather than
reusable constants.

Parameterize episode-specific values without abstracting away the formula,
threshold, relationship between inputs, requested count, ordering constraint,
selection rule, or visible verification that makes the procedure correct. Do
not promote one observed layout, value, or incidental action into a universal
requirement.

When evidence is incomplete, state an observable conditional check or supported
limitation in the Skill. Never invent a hidden rubric or success condition.

\promptheading{Write the Skill}
Create \texttt{<skill-name>/SKILL.md} with \toolname{write\_file}. Start with exactly
this frontmatter shape:

\begin{promptcode}
\texttt{---}\par
name: skill-name\par
description: Explain what the skill does and the situations that should trigger it.\par
\texttt{---}\par
\end{promptcode}

Write the body in clear imperative language. Include the complete workflow,
selection and calculation rules, scope and count constraints, phase handoffs,
visible completion checks, and only recovery guidance supported by evidence.
Avoid introductory material a capable execution agent already knows, but never
remove a step, dependency, decision, stop condition, or recovery check that can
affect correctness.

Use semantic UI targets instead of fixed coordinates. When labels or layouts
may vary, give a stable visible role or relationship that identifies the target
without guessing. For persistent changes, verify the resulting visible state;
refresh, revisit, or reopen the relevant view when the observed workflow shows
that this is necessary to distinguish a saved result from a transient one.

Add \texttt{references/} only for genuinely supporting knowledge, never as a place to
hide required runtime instructions. Add \texttt{scripts/} or \texttt{assets/} only when they
are essential to reuse and their correctness can be established from the
available evidence. Do not initialize templates, run packaging scripts, or
create archives.

\promptheading{Self-Review}
Before finalization, read every final \texttt{SKILL.md} back with \toolname{read\_file} after its
last write and verify:
\begin{promptitems}
\item the folder name and frontmatter name match, the description identifies the
right trigger, and the body is non-empty;
\item an execution agent can complete the workflow without the source recording,
trace, extraction manifest, or unstated episode knowledge;
\item graphical tasks make state changes through visible UI actions and do not
offer backend or developer shortcuts unless the task explicitly requires one;
\item every explicit goal, formula, operand, unit, quantity, scope, selection rule,
ordering relationship, phase handoff, and count-based stop condition needed
at runtime is present;
\item semantic targets replace coordinates and incidental layout;
\item critical mutations and the final result have observable UI checks, including
persistence checks where required;
\item every instructed persistent mutation is authorized by the task;
\item supported failure modes retain their recovery and stop conditions;
\item episode-specific values are parameterized without erasing the relationships
that govern their use;
\item material uncertainty is an observable conditional check or limitation, not an
invented action or hidden success criterion;
\item every bundled file is referenced, necessary, and contains no core instruction
that belongs in \texttt{SKILL.md}.
\end{promptitems}

Call \toolname{finalize\_extraction}. The runtime writes the extraction manifest. Treat
the finalizer result as authoritative: correct every reported issue and call it
again. Do not finish with an ordinary text response.
\end{promptbox}

\subsection{Downstream Execution Prompt}
\label{app:execution-prompt}
We append the following shared wrapper to the executor's base system prompt
for \ours, ShowUI-Aloha, and D2T. Only the guidance text changes between methods,
as detailed in Table~\ref{tab:injected-guidance}; No Skill adds no guidance block.
The wrapper directs the executor to adapt the guidance to the current task
and screen and to verify the result in the live environment.

\begin{promptbox}{Shared Downstream Guidance Wrapper}
\texttt{<task\_skills>}

The following reusable skills are advisory guidance for this task. Adapt them
to the live screen and current task values. The user's task instruction remains
authoritative. Do not assume recorded coordinates, timestamps, identities,
instance values, or completion state. Verify the requested result in the live
environment before finishing.

\texttt{[guidance\_text]}

\texttt{</task\_skills>}
\end{promptbox}

\begin{table}[H]
    \centering
    \small
    \setlength{\tabcolsep}{4pt}
    \renewcommand{\arraystretch}{1.15}
    \caption{Method-specific contents inserted into the shared
    \texttt{<task\_skills>} wrapper at the end of the system prompt.
    No Skill adds no block.}
    \label{tab:injected-guidance}
    \begin{tabularx}{\linewidth}{@{}>{\raggedright\arraybackslash}p{0.16\linewidth}
        >{\raggedright\arraybackslash}X >{\raggedright\arraybackslash}X@{}}
        \toprule
        Method & Retained content & Text format \\
        \midrule
        No Skill & None & No appended block. \\
        \addlinespace
        \ours & Complete task-matched \texttt{SKILL.md} files, including
        their frontmatter and instructions.
        & Each file follows a skill-directory heading. Multiple files are
        concatenated in the task map's order, separated by blank lines. \\
        \addlinespace
        ShowUI-Aloha & Each trajectory step's \texttt{caption.action} text.
        & A task heading followed by \texttt{Step 1: ...},
        \texttt{Step 2: ...}, and subsequent actions in trajectory order. \\
        \addlinespace
        D2T & Tutorial prose after removing frontmatter, image-containing
        lines, and link destinations.
        & Original section and step headings and remaining prose;
        ordinary hyperlink labels are preserved. \\
        \bottomrule
    \end{tabularx}
\end{table}

\section{Benchmark Tasks and Evaluation}
\label{app:benchmark}

\paragraph{Task screening and demonstration review.}
We first exclude infeasible tasks for which a valid human demonstration cannot
be completed. We then review the human recordings and discard demonstrations
that are incomplete, omit key operations, or fail to complete the task correctly.
Only tasks and their associated demonstrations that pass this screening and
review are retained.

\subsection{MyPCBench Tasks and Jitter Perturbations}
\label{app:jitter-settings}
We use the 64-task MyPCBench Bounded Action task set. Each task is paired
with a recorded human demonstration and its exported multimodal evidence for
skill induction.

\paragraph{Perturbation construction.}
We construct one Jitter variant for each original task, yielding 64
source--variant pairs. GPT-5.6-sol assists in generating the revised task instructions and
corresponding evaluation criteria, followed by human verification. The
perturbations modify task requirements within the original application context:
target entities and source materials; numerical and temporal parameters;
selection and ranking rules; and operation constraints or required final states.
Table~\ref{tab:jitter-examples} gives representative examples.

\begin{figure}[H]
    \centering
    \includegraphics[width=\linewidth]{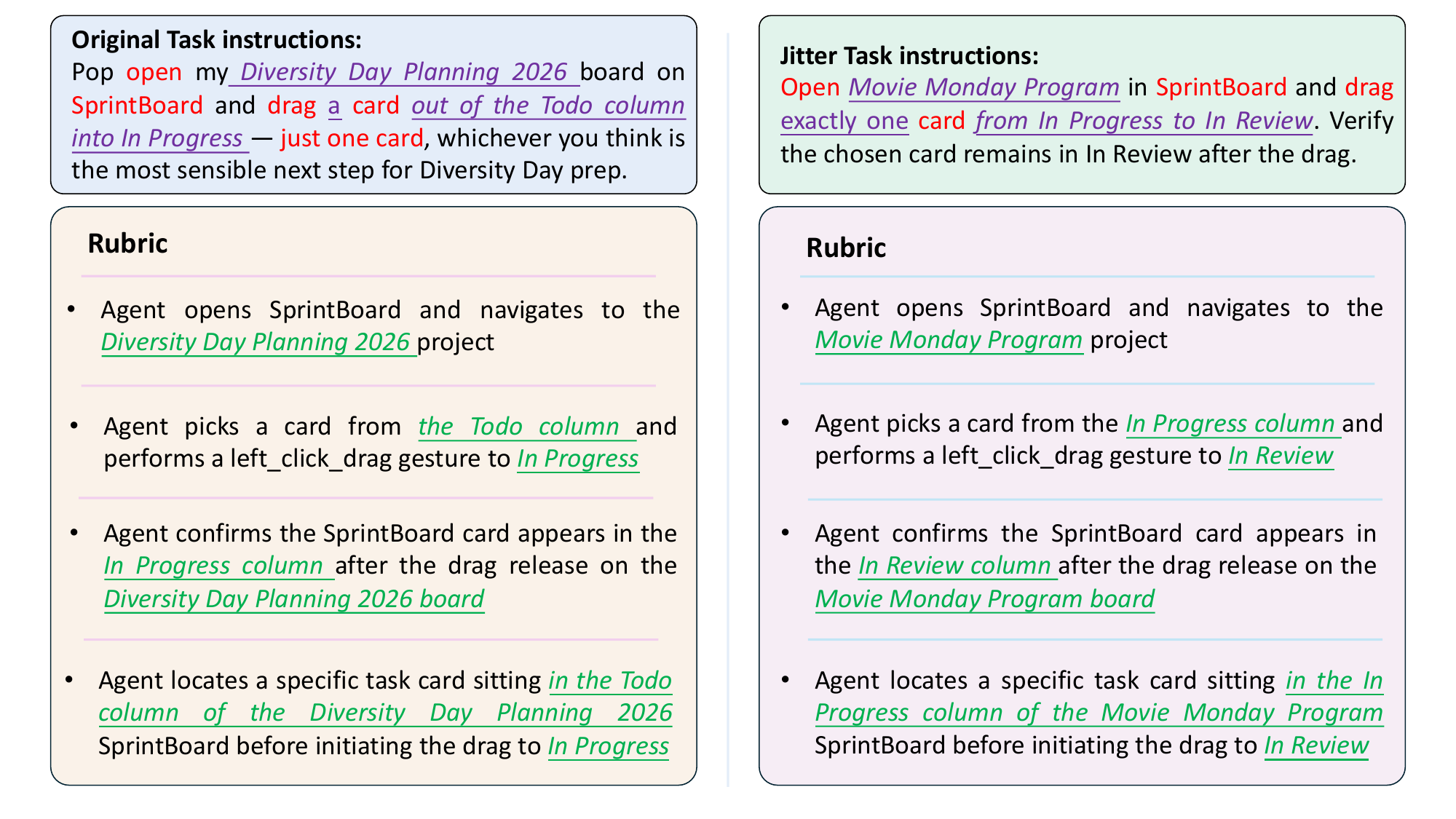}
    \caption{A paired SprintBoard task and its evaluation rubric before (left)
    and after (right) Jitter perturbation. The variant changes the target board
    and the required card transition, with rubric items revised to match the
    new requirements.}
    \label{fig:jitter-task-rubric}
\end{figure}

\paragraph{Evaluation criteria and skill reuse.}
All 64 task instructions are revised. Across the 366 paired rubric items, 331
criteria are updated to match the perturbed requirements and 35 generic criteria
are retained. All rubric weights and evaluator types remain unchanged. Each
Jitter task retains the skill-file mapping of its paired original task, so the
evaluation measures reuse of the existing skill guidance on a changed task
instance. Figure~\ref{fig:jitter-task-rubric} illustrates paired changes to a
SprintBoard task and its evaluation criteria.

\begin{table}[H]
    \centering
    \small
    \setlength{\tabcolsep}{4pt}
    \renewcommand{\arraystretch}{1.15}
    \caption{Representative changes in the paired MyPCBench Jitter tasks.
    Descriptions summarize the changed requirements.}
    \label{tab:jitter-examples}
    \begin{tabularx}{\linewidth}{@{}p{0.19\linewidth}XX@{}}
        \toprule
        Application & Original requirements & Jitter requirements \\
        \midrule
        Dinoco Airlines
        & Book an AVP--JFK flight for next Friday evening or night with a
          left-side seat.
        & Book an AVP--PHL flight for next Tuesday morning with a
          right-side seat. \\
        \addlinespace
        OddsMarket
        & Select the three largest positions by share count; break ties by
          higher cost basis, then alphabetically.
        & Select the two smallest positions by share count; break ties by
          lower cost basis, then alphabetically. \\
        \addlinespace
        SpeedTax
        & Advance the in-progress current-year return to the review step
          without submitting it.
        & Audit the filed 2024 return read-only and report refunds, confirmation
          number, and accepted status. \\
        \bottomrule
    \end{tabularx}
\end{table}

\subsection{OSWorld-Verified Task Distribution}
\begin{figure}[H]
    \centering
    \includegraphics[width=\linewidth]{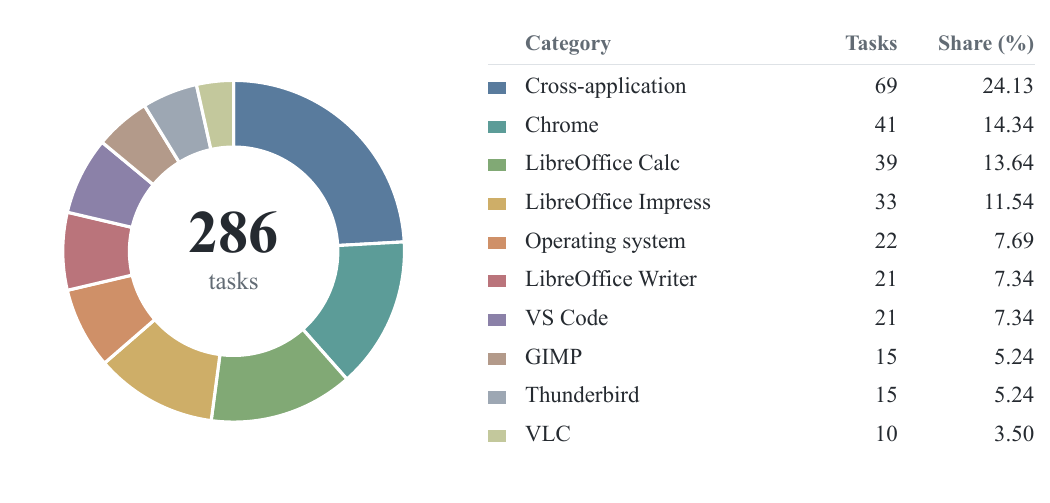}
    \caption{Category distribution of the 286 retained OSWorld-Verified tasks.
    The legend gives the task count and percentage for each category.}
    \label{fig:osworld-task-distribution}
\end{figure}

The curated OSWorld-Verified subset contains 286 tasks across ten categories.
Figure~\ref{fig:osworld-task-distribution} summarizes the category distribution.
Percentages are calculated over the 286 retained tasks.
The complete task-ID list appears in Table~\ref{tab:osworld-task-ids}
in Appendix~\ref{app:osworld-task-ids}.

\subsection{Execution Harness and Evaluator Configuration}
\label{app:evaluator-config}
\paragraph{OSWorld execution harness.}
We use a tool-calling desktop agent connected to OSWorld's official
\texttt{DesktopEnv}. Each task is initialized from its benchmark configuration.
Given the task instruction, the agent uses the current screenshot and action
history to select its next GUI actions, with guidance appended as described in
Appendix~\ref{app:execution-prompt}. Its tools support screenshots, pointer
movement, clicks, drags, scrolling, text entry, key presses, and waiting.
A desktop adapter maps screenshot-relative pixel coordinates to the VM and
executes GUI actions through PyAutoGUI, returning post-action screenshots and
textual results. Each execution step is one model iteration and may contain
multiple tool calls. The loop stops on a terminal decision or at the step limit
specified in Appendix~\ref{app:model-settings}.

\paragraph{Evaluation and metrics.}
For MyPCBench, we use the official environment setup and evaluator. For
OSWorld-Verified, we reuse the official
task-specific evaluators to score the resulting environment states. Evaluators run
after the agent terminates or reaches the maximum action budget. For partial
MyPCBench rubrics, the task score is the fraction or weighted fraction of
satisfied items according to the official implementation. Full score requires
all rubric conditions. OSWorld-Verified evaluators can return partial scores.
For both benchmarks, Score averages the evaluator outputs and scales the result
to 0--100, while SR counts only tasks receiving full credit. This SR corresponds
to MyPCBench's perfect rate; OSWorld's official mean-score aggregation, called
``success rate'' in its reporting script, is reported as Score in our tables.
We store evaluator logs and final-state artifacts for every rollout.

Let \(g_i \in [0,1]\) denote the normalized evaluator output for task
\(i\), where \(1\) indicates full credit. For a task set
\(\mathcal{D}\), we compute the mean Score and full-credit success rate
SR (\%) as follows:
\begin{equation}
\mathrm{Score}
=
\frac{100}{|\mathcal{D}|}
\sum_{i \in \mathcal{D}} g_i,
\qquad
\mathrm{SR}
=
\frac{100}{|\mathcal{D}|}
\sum_{i \in \mathcal{D}}
\mathbf{1}\!\left[g_i = 1\right].
\end{equation}

\subsection{Baseline Adaptation for Fair Comparison}
\label{app:baseline-adaptation}
No Skill (also labeled Baseline) provides no additional guidance.
We evaluate ShowUI-Aloha and D2T as sources of guidance for the same downstream
execution agent. For ShowUI-Aloha, we retain each trajectory step's
\texttt{caption.action} field and concatenate the actions in their original
order; observation, reasoning, and expectation fields are excluded. For D2T,
we read the generated tutorial's \texttt{index.md}, remove frontmatter and
image-containing lines, and strip link destinations and bare URLs while
preserving ordinary hyperlink labels. The remaining tutorial headings and
prose retain their original order. All retained text is supplied through the
same wrapper as our skills (Appendix~\ref{app:execution-prompt}), with the
method-specific contents summarized in Table~\ref{tab:injected-guidance}.

For a given execution model, all methods share the base prompt, observations,
action interface, step limit, termination policy, and evaluator. On MyPCBench,
this shared action interface includes both GUI actions and Bash commands.
When their required modalities are available, they use the same source demonstration.
The reported results measure how effectively these adapted guidance outputs
support task execution under the same downstream conditions.

\subsection{Model Pairing and Runtime Settings}
\label{app:model-settings}
The default pairing uses the same model for skill induction and downstream
execution: each of the four execution models uses its own induced skills.
The cross-model experiment explicitly varies this pairing
(Section~\ref{sec:cross-model}). Tables~\ref{tab:model-config}
and~\ref{tab:execution-config} report our settings for skill induction and
downstream execution, respectively. The induction limit counts agent iterations; the
execution limit counts steps. The execution history and screenshot windows
are configured separately from token limits.

\begin{table}[H]
    \centering
    \begin{minipage}[t]{0.48\linewidth}
    \centering
    \caption{Our skill-induction settings.}
    \label{tab:model-config}
    \small
    \begin{tabularx}{\linewidth}{@{}Xr@{}}
        \toprule
        Setting & Value \\
        \midrule
        Reasoning effort & high \\
        Temperature & 0 \\
        Context limit (tokens) & 200,000 \\
        Output limit per model call (tokens) & 16,000 \\
        Induction iteration limit & 100 \\
        \bottomrule
    \end{tabularx}
    \end{minipage}\hfill
    \begin{minipage}[t]{0.48\linewidth}
    \centering
    \caption{Our downstream-execution settings.}
    \label{tab:execution-config}
    \small
    \begin{tabularx}{\linewidth}{@{}Xr@{}}
        \toprule
        Setting & Value \\
        \midrule
        Reasoning effort & high \\
        Temperature & 0 \\
        Output limit per model call (tokens) & 16,000 \\
        History window (turns) & 100 \\
        Screenshot window (images) & 3 \\
        Execution step limit & 100 \\
        \bottomrule
    \end{tabularx}
    \end{minipage}
\end{table}

\paragraph{Execution context management.}
When a downstream model request exceeds the context limit, the system retains
the task instruction, skill guidance, and most recent screenshot, clears the
historical interaction messages, and retries the request. This shortens the
context while preserving the task requirements and visual evidence of the
current interface.

\subsection{Evidence-Modality Ablation Settings}
\label{app:modality-settings}
The ablations in Section~\ref{sec:induction-ablation} vary the recording
evidence available to the induction agent. The task instruction remains
available in every condition. Table~\ref{tab:modality-settings} lists the
retained evidence and visual tools. We remove only tools that have no applicable
input under a condition. File discovery, text reading, skill authoring, and
finalization use the same tools described in Appendix~\ref{app:tools}.

\begin{table}[H]
    \centering
    \small
    \setlength{\tabcolsep}{4pt}
    \renewcommand{\arraystretch}{1.15}
    \caption{Evidence and visual tools available for skill induction in each
    modality condition. Task instructions and common file-reading and
    skill-authoring tools remain available. UI annotations denote metadata and
    descriptions derived from each action's target element in the AX tree.}
    \label{tab:modality-settings}
    \begin{tabularx}{\linewidth}{@{}p{0.19\linewidth}X p{0.19\linewidth}@{}}
        \toprule
        Setting & Retained recording evidence & Visual tools \\
        \midrule
        Full Evidence & Actions with UI annotations, KeyLogs, indexed images, video
        & Image and video \\
        \addlinespace
        w/o Indexed Images & Actions with UI annotations, KeyLogs, video & Video \\
        \addlinespace
        w/o UI Annotations & Actions with generic captions, KeyLogs, images with UI target annotations removed, video
        & Image and video \\
        \addlinespace
        w/o KeyLogs & Actions with UI annotations, event frames without added overlays, video
        & Image and video \\
        \addlinespace
        Video-only & Raw video & Video \\
        \addlinespace
        UI-Grounded Action Text Only & Ordered text describing recorded actions and their target UI elements & None \\
        \bottomrule
    \end{tabularx}
\end{table}

In Table~\ref{tab:modality-settings}, the image tool is
\toolname{read\_visual\_evidence}; video tools are
\toolname{get\_video\_metadata}, \toolname{watch\_video}, and
\toolname{extract\_video\_frames}. Removing UI annotations or KeyLogs does not
require removing \toolname{read\_file}, which still serves the remaining
textual evidence and task instruction.

\paragraph{Image and video access.}
\textbf{w/o Indexed Images} removes the exported \texttt{evidence/} directory and the
image-reading tool, while retaining the video and all video tools. The agent
can therefore still inspect video frames; this condition tests the contribution
of indexed, event-aligned images when video remains available.
\textbf{Video-only} exposes only \texttt{task.json} and \texttt{video.mp4}.
Video metadata and frame timestamps remain available through the video tools;
no action trace or precomputed image index is supplied.

\paragraph{Removing derived annotations.}
The condition labeled \textbf{w/o UI Annotations} corresponds to the implementation's
\texttt{without\_dom} protocol. It removes AX-derived target-element fields from
actions and replaces the corresponding target descriptions with generic action
captions. Image indices are sanitized, and the top annotation bar is replaced
with raw-video pixels from the same timestamp, preserving click and drag
highlights.

\textbf{w/o KeyLogs} removes \texttt{trace/keylog.jsonl} and keylog-derived
image metadata, including input-event labels, raw messages, and source-image
references. Indexed frames are re-extracted from raw video at their original
timestamps, removing all added image overlays, including the top caption bar
and click/drag highlights. The action trace and its AX-derived target
information remain unchanged, as do the action captions in the image index.
Thus, this condition measures the joint effect of removing the raw log and
the annotated image presentation; it does not isolate the log file's
contribution or remove all keyboard information from the retained actions.

\paragraph{UI-grounded action text only.}
This condition retains only the task instruction and ordered action captions in
\texttt{task.json} and \texttt{trace/actions.jsonl}, respectively. The captions combine recorded
actions with available target UI descriptions, such as ``click the Search
button.'' Target descriptions can include control names and roles from the
corresponding target elements in the AX tree. Each action row retains only its original
\texttt{caption}; coordinates, timestamps, action IDs, structured target metadata,
and other action metadata are removed. Raw keylogs, images, and video are
absent, and all image and video tools are disabled. Thus, UI target semantics
are preserved in text rather than supplied as a full AX Tree. This condition
evaluates skill induction from these descriptions; ShowUI-Aloha instead
supplies action semantics directly as downstream guidance.

\paragraph{Prompt adaptation.}
The five ablations share a prompt instructing the agent to inspect the available
evidence, followed by a condition-specific statement listing the exposed
modalities. This adapts the full-evidence prompt's requirement to inspect
representative initial, critical, and final-state images to the available inputs.
Task-first reading, the skill-authoring policy, output validation, and
finalization requirements are retained.

\Needspace{12\baselineskip}
\section{Case Study and Skill Induction Example}
\label{app:examples}

\subsection{Task and Induction Trace}
We present the generated skill for MyPCBench Bounded Action task
\texttt{hard\_app-f016}, a TableFind reservation task also used in the main-text
case study (Section~\ref{sec:case-study}). The task instruction is:

\begin{quote}
Book me Cotogna for ten people for the Dundies after-party next Friday.
While you're in TableFind, pull up three other Scranton-area restaurants as
a fallback plan in case Cotogna is full, and favorite the top two. Once that's
done, confirm the Cotogna reservation with a note that it's the Dundies
after-party so they can prep the vibe.
\end{quote}

The recorded induction run uses GPT-5.5 with high reasoning effort and makes
30 tool calls. It reads the task, episode metadata, summary, evidence indices,
action trace, and key log; inspects 22 distinct evidence images in two batches
of 14 and eight; and samples the 156.47-second video in two calls to
\toolname{watch\_video}, after reading its metadata. The video calls return
13 and eight frames, respectively, with one shared timestamp. The agent writes
a draft, reads it back, revises it, and reads the final skill before successful
\toolname{finalize\_extraction}. This record documents the induction output;
we do not assign a downstream execution score to this listing.

\subsection{Detailed Execution Analysis}
\label{app:case-analysis}
Figure~\ref{fig:case-study} shows GPT-5.5's execution trajectories for the case
in Section~\ref{sec:case-study}. Table~\ref{tab:case-rubrics} reports all eight
rubric outcomes under ShowUI-Aloha, D2T, and \ours guidance.

\paragraph{Task dependencies.}
The instruction combines a primary reservation with a fallback workflow:
inspect three other Scranton-area restaurants and favorite the top two before
confirming Cotogna with the requested note. Completing this workflow requires
carrying the date and party size across searches, distinguishing the primary
restaurant from alternatives, and returning to the primary booking after
handling the fallbacks. The evaluation checks both the reservation itself
and these additional requirements, including persistence after a refresh.

\begin{table}[t]
    \centering
    \small
    \setlength{\tabcolsep}{4pt}
    \renewcommand{\arraystretch}{1.15}
    \caption{Rubric-level outcomes for the TableFind case in
    Figure~\ref{fig:case-study}. Requirement descriptions summarize R1--R8;
    Aloha denotes the adapted ShowUI-Aloha baseline. Scores are the reported
    evaluator outputs.}
    \label{tab:case-rubrics}
    \begin{tabularx}{\linewidth}{@{}lXccc@{}}
        \toprule
        Item & Requirement & Aloha & D2T & \ours \\
        \midrule
        R1 & Set the requested date and a table for ten. & Pass & Pass & Pass \\
        R2 & Book Cotogna with the requested note. & Pass & Pass & Pass \\
        R3 & Open three alternative restaurants. & Fail & Fail & Pass \\
        R4 & View each alternative's detail page. & Fail & Fail & Pass \\
        R5 & Save the top two alternatives. & Fail & Fail & Pass \\
        R6 & Obtain a reservation confirmation number. & Pass & Pass & Pass \\
        R7 & Check persistence after a refresh. & Fail & Fail & Pass \\
        R8 & Verify the correct final details. & Pass & Pass & Pass \\
        \midrule
        \multicolumn{2}{l}{Evaluator score} & \(55/100\) & \(55/100\) & \(100/100\) \\
        \bottomrule
    \end{tabularx}
\end{table}

\paragraph{Where the baselines fall short.}
Both baselines pass R1, R2, R6, and R8: they configure the booking,
include the note, obtain confirmation, and verify the final reservation
details. Their trajectories reach alternative-restaurant views, but the
rubric outcomes do not establish that all three alternatives and their
detail pages were inspected (R3--R4). Saving the top two alternatives
is also incomplete (R5). Both fail the separate persistence check (R7),
despite passing the check of final details (R8). A visible confirmation
and a correct reservation display therefore cover only part of the
required workflow. Each baseline receives \(55/100\).

\Needspace{7\baselineskip}
\paragraph{What the successful trajectory preserves.}
\ours passes all eight items and receives \(100/100\). Its four additional
passed criteria concern alternative-restaurant coverage, saving the preferred
options, and persistence after refresh. Figure~\ref{fig:case-study} shows
these steps alongside the primary booking and confirmation. The comparison
pinpoints the benefit in this case: completing the secondary subgoals and
verification requirements that remain unfinished in the baseline trajectories.

\paragraph{Connection to the generated procedure.}
The full skill in Appendix~\ref{app:tablefind-skill} makes several relevant
dependencies explicit. It carries booking values across searches, rechecks
fields that may reset, excludes the primary restaurant from the fallback
set, and selects the first two qualifying alternatives under the active
ranking. It also defers final confirmation until the fallback steps are
complete and checks both the reservation details and the saved favorites.
These instructions organize the task into dependent stages with observable
completion checks and explicit reminders of the secondary requirements
across the primary reservation and fallback searches.

\subsection{Complete Generated Skill}
\label{app:tablefind-skill}
The skill below is reproduced in full, with only Markdown presentation adapted
to the appendix format. It carries the date, time, party size, and note across
the primary and fallback searches; checks for search-field resets; excludes
the primary restaurant from the fallback set; and preserves the ordering of
fallback selection, favoriting, and final confirmation. Restaurant names and
booking values are taken from the task, while the generated procedure retains
the Scranton-area scope. Its final checks use visible reservation and favorite
states.

\clearpage
\begin{promptbox}{Generated Skill: TableFind Reservation with Fallbacks}
\begin{promptcode}
---\par
name: tablefind-reservation-with-fallbacks\par
description: Use TableFind to reserve a requested restaurant for a specified party/date/time, identify fallback Scranton-area restaurants, favorite the requested top fallback choices, and verify the confirmed reservation details.\par
---\par
\end{promptcode}

Use this workflow when a task asks for a TableFind restaurant reservation plus fallback restaurants and favorites.

\promptheading{Inputs to carry through the workflow}
\begin{promptitems}[itemsep=3pt]
\item Primary restaurant name.
\item Party size.
\item Reservation date expression or exact date. Resolve relative dates such as "next Friday" before interacting with the calendar; if the task is based on the current TableFind/session date, use that context to choose the next matching weekday.
\item Reservation time. If the task does not specify a time and TableFind already provides a default time in the search form, leave the default time unchanged rather than inventing a different one.
\item Reservation note or special-request text.
\item Fallback scope, count, and favorite rule, such as "three other Scranton-area restaurants" and "favorite the top two."
\end{promptitems}

Keep the same date, time, and party size for fallback searches unless the task explicitly gives different fallback criteria. Fallback restaurants must be different from the primary restaurant.

\promptheading{Set the primary search criteria}
\begin{promptsteps}[itemsep=3pt]
\item Open TableFind and use the visible Search page or TableFind logo/home link to reach the main search form.
\item Set the Date field to the task date. Use the calendar picker and select the resolved day; verify the date field updates to the intended date.
\item Leave or set the Time field to the task time. If no time is specified, keep TableFind's visible default time.
\item Open the Party Size combo box and choose the requested party size, e.g. the menu item labeled with the required number of people/guests. Verify the party-size field shows the requested count.
\item Click the Restaurant or Cuisine entry, type the primary restaurant name, and click Search.
\item On the results page, verify the heading matches the requested party size, date, and time, and select the result whose visible name exactly matches the primary restaurant.
\end{promptsteps}

\promptheading{Check and prepare the primary reservation without losing task details}
\begin{promptsteps}[itemsep=3pt]
\item On the primary restaurant detail page, verify the restaurant name, cuisine/location summary, booking panel date, party size, and available time slots are visible.
\item Select the requested time slot if it is not already selected. The selected summary should show the intended date and time and indicate remaining tables, not sold out or waitlist. If the exact requested/default time is unavailable, do not confirm an alternate unless the task authorizes alternatives.
\item Enter the task's note in the Special requests field exactly enough to communicate the requested purpose or constraint. Verify the note text is visible in the field.
\item If the task says to gather/favorite fallbacks before confirming the primary reservation, do not click the final Confirm Reservation yet. You may leave the primary detail page and return later, carrying forward the restaurant name, date, time, party size, and note.
\end{promptsteps}

\promptheading{Pull up fallback restaurants and favorite the top choices}
\begin{promptsteps}[itemsep=3pt]
\item Navigate back to the TableFind Search/results view using the TableFind logo, Search link, or Back to search control.
\item Clear the primary restaurant-name search if necessary so that the page shows general Scranton-area restaurant results for the same date, time, and party size. Keep or choose a ranking such as Sort: Top Rated when the task asks for "top" restaurants and no other ranking is specified.
\item Verify the results heading still reflects the intended date, time, and party size. If the page has reverted to default values, reset the date, time, and party size before relying on the fallback list.
\item Starting from the top of the ranked Scranton-area results, identify fallback restaurants that are not the primary restaurant. Count only distinct restaurant cards/pages with Scranton-area neighborhood or address information.
\item Pull up the requested number of fallback restaurants by opening each qualifying restaurant's detail page or otherwise bringing its card/details into view. For each one, confirm it is a different Scranton-area restaurant and that it has availability for the carried date/time/party criteria or a visible set of usable time slots for those criteria. Return to the results list after inspecting each fallback.
\item Favorite the requested top count according to the active ranking. For "favorite the top two," favorite the first two qualifying fallback restaurants in the ranked list, not arbitrary later restaurants. Use each restaurant card's heart / Save to favorites toggle.
\item After each favorite action, verify visible confirmation: the heart becomes filled/active or the card text changes to "Saved to Favorites." Do not toggle an already-saved heart off; if it is already active, count it as favorited and leave it active.
\end{promptsteps}

\promptheading{Confirm the primary reservation}
\begin{promptsteps}[itemsep=3pt]
\item Return to the primary restaurant detail page using Search, the primary restaurant name, browser history, or a saved/open page. Re-check the restaurant name, date, time, party size, and special-request note.
\item Click Complete reservation only when the booking panel shows the intended selected date/time and party size and the special-request note is present.
\item In the confirmation dialog, compare every visible row with the task requirements:
\begin{promptitems}[itemsep=3pt]
\item Restaurant: primary restaurant name.
\item Date \& time: resolved task date and intended time.
\item Party size: requested number of guests/people.
\item Special requests: the requested note text.
\end{promptitems}
\item If any row is wrong, cancel the dialog and correct the visible fields before trying again. If all rows are correct, click Confirm Reservation.
\end{promptsteps}

\promptheading{Final verification}
\begin{promptsteps}[itemsep=3pt]
\item Open My Reservations after confirming.
\item Find the newly created primary reservation. Verify it is marked Confirmed and shows the correct restaurant, date/time, party size, and special-request note. A confirmation code may also be present, but the visible restaurant/date/party/note/status are the essential checks.
\item Verify the requested number of fallback favorites remain active. If still on the results page, check the filled hearts or "Saved to Favorites" labels for the top fallback restaurants. If needed, open the Saved page from the header and confirm those fallback restaurants appear there.
\item Stop only after the primary reservation is confirmed and the required top fallback restaurants are visibly favorited.
\end{promptsteps}
\end{promptbox}

\clearpage
\section{Tool Usage During Skill Induction}
\label{app:efficiency}
\label{app:induction-tool-usage}
\begin{figure}[H]
    \centering
    \includegraphics[width=\linewidth]{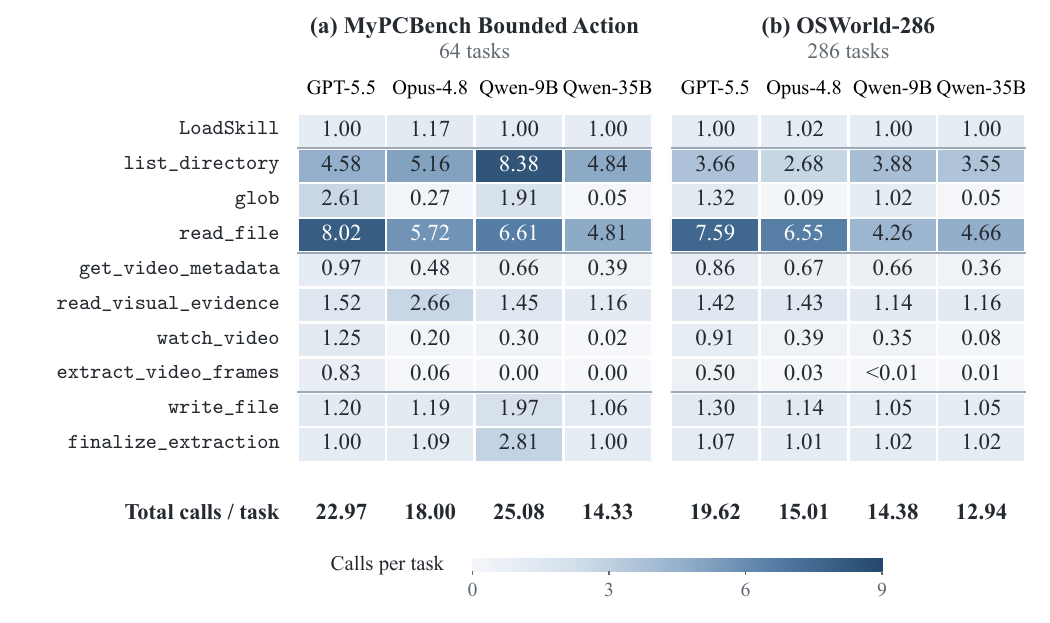}
    \caption{Mean tool calls per task during skill induction, using a shared
    linear color scale. Values are rounded to two decimals; positive values
    that would round to zero are shown as $<0.01$. Horizontal rules separate
    tool groups; the bold row reports mean total calls per task. Qwen-9B and
    Qwen-35B denote Qwen3.5-9B and Qwen3.5-35B-A3B, respectively.}
    \label{fig:induction-tool-usage}
\end{figure}

Figure~\ref{fig:induction-tool-usage} summarizes tool usage during skill induction
by four models on 64 MyPCBench Bounded Action tasks and the 286-task
OSWorld-Verified subset. Each value is the aggregate invocation count divided
by the dataset's task count; the summary row includes all ten tools.

GPT-5.5 invokes \toolname{watch\_video} and
\toolname{extract\_video\_frames} most frequently on both datasets, while
Qwen3.5-35B-A3B makes the fewest total calls per task (14.33 and 12.94).
Invocation counts alone do not establish evidence coverage or skill quality:
a call can return multiple images or varying amounts of text. Repeated
\toolname{finalize\_extraction} calls also do not by themselves identify failed
induction runs. When validation fails, the tool returns diagnostics so the
agent can address the reported issues and retry. The run can still complete
successfully after these intermediate attempts.

\clearpage
\section{Complete OSWorld-Verified Task List}
\label{app:osworld-task-ids}
Table~\ref{tab:osworld-task-ids} lists all 286 OSWorld-Verified tasks used
for demonstration collection and evaluation. Task IDs are preserved in full
and ordered lexicographically, read left to right across each row.

\begin{table}[H]
    \centering
    \small
    \caption{Complete task IDs for the 286-task OSWorld-Verified subset.}
    \label{tab:osworld-task-ids}
    \renewcommand{\arraystretch}{1.07}
    \begin{tabular}{@{}r@{\hspace{5pt}}>{\ttfamily}l@{\hspace{16pt}}r@{\hspace{5pt}}>{\ttfamily}l@{}}
        \toprule
        No. & \multicolumn{1}{l}{Task ID} & No. & \multicolumn{1}{l}{Task ID} \\
        \midrule
        1 & 00fa164e-2612-4439-992e-157d019a8436 & 2 & 01b269ae-2111-4a07-81fd-3fcd711993b0 \\
        3 & 02ce9a50-7af2-47ed-8596-af0c230501f8 & 4 & 030eeff7-b492-4218-b312-701ec99ee0cc \\
        5 & 0326d92d-d218-48a8-9ca1-981cd6d064c7 & 6 & 035f41ba-6653-43ab-aa63-c86d449d62e5 \\
        7 & 04d9aeaf-7bed-4024-bedb-e10e6f00eb7f & 8 & 0512bb38-d531-4acf-9e7e-0add90816068 \\
        9 & 06ca5602-62ca-47f6-ad4f-da151cde54cc & 10 & 06fe7178-4491-4589-810f-2e2bc9502122 \\
        11 & 0810415c-bde4-4443-9047-d5f70165a697 & 12 & 08aced46-45a2-48d7-993b-ed3fb5b32302 \\
        13 & 08c73485-7c6d-4681-999d-919f5c32dcfa & 14 & 0a0faba3-5580-44df-965d-f562a99b291c \\
        15 & 0a211154-fda0-48d0-9274-eaac4ce5486d & 16 & 0a2e43bf-b26c-4631-a966-af9dfa12c9e5 \\
        17 & 0b17a146-2934-46c7-8727-73ff6b6483e8 & 18 & 0bf05a7d-b28b-44d2-955a-50b41e24012a \\
        19 & 0cecd4f3-74de-457b-ba94-29ad6b5dafb6 & 20 & 0d8b7de3-e8de-4d86-b9fd-dd2dce58a217 \\
        21 & 0e47de2a-32e0-456c-a366-8c607ef7a9d2 & 22 & 0e763496-b6bb-4508-a427-fad0b6c3e195 \\
        23 & 0ed39f63-6049-43d4-ba4d-5fa2fe04a951 & 24 & 0f84bef9-9790-432e-92b7-eece357603fb \\
        25 & 10a730d5-d414-4b40-b479-684bed1ae522 & 26 & 12086550-11c0-466b-b367-1d9e75b3910e \\
        27 & 121ba48f-9e17-48ce-9bc6-a4fb17a7ebba & 28 & 1273e544-688f-496b-8d89-3e0f40aa0606 \\
        29 & 1334ca3e-f9e3-4db8-9ca7-b4c653be7d17 & 30 & 13584542-872b-42d8-b299-866967b5c3ef \\
        31 & 15aece23-a215-4579-91b4-69eec72e18da & 32 & 15c3b339-88f7-4a86-ab16-e71c58dcb01e \\
        33 & 1704f00f-79e6-43a7-961b-cedd3724d5fd & 34 & 1954cced-e748-45c4-9c26-9855b97fbc5e \\
        35 & 1d17d234-e39d-4ed7-b46f-4417922a4e7c & 36 & 1de60575-bb6e-4c3d-9e6a-2fa699f9f197 \\
        37 & 1e8df695-bd1b-45b3-b557-e7d599cf7597 & 38 & 1f18aa87-af6f-41ef-9853-cdb8f32ebdea \\
        39 & 20236825-b5df-46e7-89bf-62e1d640a897 & 40 & 215dfd39-f493-4bc3-a027-8a97d72c61bf \\
        41 & 21760ecb-8f62-40d2-8d85-0cee5725cb72 & 42 & 21ab7b40-77c2-4ae6-8321-e00d3a086c73 \\
        43 & 227d2f97-562b-4ccb-ae47-a5ec9e142fbb & 44 & 23393935-50c7-4a86-aeea-2b78fd089c5c \\
        45 & 236833a3-5704-47fc-888c-4f298f09f799 & 46 & 2373b66a-092d-44cb-bfd7-82e86e7a3b4d \\
        47 & 26150609-0da3-4a7d-8868-0faf9c5f01bb & 48 & 26660ad1-6ebb-4f59-8cba-a8432dfe8d38 \\
        49 & 26a8440e-c166-4c50-aef4-bfb77314b46b & 50 & 276cc624-87ea-4f08-ab93-f770e3790175 \\
        51 & 2888b4e6-5b47-4b57-8bf5-c73827890774 & 52 & 28cc3b7e-b194-4bc9-8353-d04c0f4d56d2 \\
        53 & 2ad9387a-65d8-4e33-ad5b-7580065a27ca & 54 & 2ae9ba84-3a0d-4d4c-8338-3a1478dc5fe3 \\
        55 & 2b9493d7-49b8-493a-a71b-56cd1f4d6908 & 56 & 2b94c692-6abb-48ae-ab0b-b3e8a19cb340 \\
        57 & 2c9fc0de-3ee7-45e1-a5df-c86206ad78b5 & 58 & 2cd43775-7085-45d8-89fa-9e35c0a915cf \\
        59 & 2fe4b718-3bd7-46ec-bdce-b184f5653624 & 60 & 3299584d-8f11-4457-bf4c-ce98f7600250 \\
        61 & 337d318b-aa07-4f4f-b763-89d9a2dd013f & 62 & 347ef137-7eeb-4c80-a3bb-0951f26a8aff \\
        63 & 35253b65-1c19-4304-8aa4-6884b8218fc0 & 64 & 357ef137-7eeb-4c80-a3bb-0951f26a8aff \\
        65 & 36037439-2044-4b50-b9d1-875b5a332143 & 66 & 3680a5ee-6870-426a-a997-eba929a0d25c \\
        67 & 368d9ba4-203c-40c1-9fa3-da2f1430ce63 & 68 & 3720f614-37fd-4d04-8a6b-76f54f8c222d \\
        69 & 37608790-6147-45d0-9f20-1137bb35703d & 70 & 37887e8c-da15-4192-923c-08fa390a176d \\
        71 & 386dbd0e-0241-4a0a-b6a2-6704fba26b1c & 72 & 39be0d19-634d-4475-8768-09c130f5425d \\
        73 & 3a7c8185-25c1-4941-bd7b-96e823c9f21f & 74 & 3a93cae4-ad3e-403e-8c12-65303b271818 \\
        75 & 3aaa4e37-dc91-482e-99af-132a612d40f3 & 76 & 3c8f201a-009d-4bbe-8b65-a6f8b35bb57f \\
        77 & 3ce045a0-877b-42aa-8d2c-b4a863336ab8 & 78 & 3e3fc409-bff3-4905-bf16-c968eee3f807 \\
        79 & 3ef2b351-8a84-4ff2-8724-d86eae9b842e & 80 & 3f05f3b9-29ba-4b6b-95aa-2204697ffc06 \\
        81 & 3f28fe4f-5d9d-4994-a456-efd78cfae1a3 & 82 & 3f49d2cc-f400-4e7d-90cc-9b18e401cc31 \\
        83 & 4127319a-8b79-4410-b58a-7a151e15f3d7 & 84 & 415ef462-bed3-493a-ac36-ca8c6d23bf1b \\
        85 & 4172ea6e-6b77-4edb-a9cc-c0014bd1603b & 86 & 4188d3a4-077d-46b7-9c86-23e1a036f6c1 \\
        87 & 42d25c08-fb87-4927-8b65-93631280a26f & 88 & 42e0a640-4f19-4b28-973d-729602b5a4a7 \\
        89 & 44ee5668-ecd5-4366-a6ce-c1c9b8d4e938 & 90 & 455d3c66-7dc6-4537-a39a-36d3e9119df7 \\
        91 & 47543840-672a-467d-80df-8f7c3b9788c9 & 92 & 47f7c0ce-a5fb-4100-a5e6-65cd0e7429e5 \\
        93 & 480bcfea-d68f-4aaa-a0a9-2589ef319381 & 94 & 48c46dc7-fe04-4505-ade7-723cba1aa6f6 \\
        95 & 48d05431-6cd5-4e76-82eb-12b60d823f7d & 96 & 4bcb1253-a636-4df4-8cb0-a35c04dfef31 \\
        \bottomrule
    \end{tabular}
\end{table}

\clearpage
\begin{table}[H]
    \centering
    \small
    \ContinuedFloat
    \caption{Complete task IDs for the 286-task OSWorld-Verified subset (continued).}
    \renewcommand{\arraystretch}{1.07}
    \begin{tabular}{@{}r@{\hspace{5pt}}>{\ttfamily}l@{\hspace{16pt}}r@{\hspace{5pt}}>{\ttfamily}l@{}}
        \toprule
        No. & \multicolumn{1}{l}{Task ID} & No. & \multicolumn{1}{l}{Task ID} \\
        \midrule
        97 & 4c26e3f3-3a14-4d86-b44a-d3cedebbb487 & 98 & 4d117223-a354-47fb-8b45-62ab1390a95f \\
        99 & 4e60007a-f5be-4bfc-9723-c39affa0a6d3 & 100 & 4ed5abd0-8b5d-47bd-839f-cacfa15ca37a \\
        101 & 4f07fbe9-70de-4927-a4d5-bb28bc12c52c & 102 & 510f64c8-9bcc-4be1-8d30-638705850618 \\
        103 & 51719eea-10bc-4246-a428-ac7c433dd4b3 & 104 & 51b11269-2ca8-4b2a-9163-f21758420e78 \\
        105 & 5203d847-2572-4150-912a-03f062254390 & 106 & 535364ea-05bd-46ea-9937-9f55c68507e8 \\
        107 & 53ad5833-3455-407b-bbc6-45b4c79ab8fb & 108 & 550ce7e7-747b-495f-b122-acdc4d0b8e54 \\
        109 & 554785e9-4523-4e7a-b8e1-8016f565f56a & 110 & 57242fad-77ca-454f-b71b-f187181a9f23 \\
        111 & 57667013-ea97-417c-9dce-2713091e6e2a & 112 & 5812b315-e7bd-4265-b51f-863c02174c28 \\
        113 & 58565672-7bfe-48ab-b828-db349231de6b & 114 & 59155008-fe71-45ec-8a8f-dc35497b6aa8 \\
        115 & 5990457f-2adb-467b-a4af-5c857c92d762 & 116 & 5ac2891a-eacd-4954-b339-98abba077adb \\
        117 & 5bc63fb9-276a-4439-a7c1-9dc76401737f & 118 & 5c1075ca-bb34-46a3-a7a0-029bd7463e79 \\
        119 & 5ca86c6f-f317-49d8-b6a7-b527541caae8 & 120 & 5ced85fc-fa1a-4217-95fd-0fb530545ce2 \\
        121 & 5cfb9197-e72b-454b-900e-c06b0c802b40 & 122 & 5d901039-a89c-4bfb-967b-bf66f4df075e \\
        123 & 5df7b33a-9f77-4101-823e-02f863e1c1ae & 124 & 5e2d93d8-8ad0-4435-b150-1692aacaa994 \\
        125 & 5ea617a3-0e86-4ba6-aab2-dac9aa2e8d57 & 126 & 6054afcb-5bab-4702-90a0-b259b5d3217c \\
        127 & 62f7fd55-0687-4a43-b6e1-3eda16fc6252 & 128 & 66399b0d-8fda-4618-95c4-bfc6191617e9 \\
        129 & 6766f2b8-8a72-417f-a9e5-56fcaa735837 & 130 & 67890eb6-6ce5-4c00-9e3d-fb4972699b06 \\
        131 & 68a25bd4-59c7-4f4d-975e-da0c8509c848 & 132 & 6a33f9b9-0a56-4844-9c3f-96ec3ffb3ba2 \\
        133 & 6ada715d-3aae-4a32-a6a7-429b2e43fb93 & 134 & 6c4c23a1-42a4-43cc-9db1-2f86ff3738cc \\
        135 & 6e99a1ad-07d2-4b66-a1ce-ece6d99c20a5 & 136 & 6f4073b8-d8ea-4ade-8a18-c5d1d5d5aa9a \\
        137 & 6f56bf42-85b8-4fbb-8e06-6c44960184ba & 138 & 6f81754e-285d-4ce0-b59e-af7edb02d108 \\
        139 & 70745df8-f2f5-42bd-8074-fbc10334fcc5 & 140 & 72b810ef-4156-4d09-8f08-a0cf57e7cefe \\
        141 & 72f83cdc-bf76-4531-9a1b-eb893a13f8aa & 142 & 734d6579-c07d-47a8-9ae2-13339795476b \\
        143 & 73c99fb9-f828-43ce-b87a-01dc07faa224 & 144 & 74d5859f-ed66-4d3e-aa0e-93d7a592ce41 \\
        145 & 7767eef2-56a3-4cea-8c9f-48c070c7d65b & 146 & 778efd0a-153f-4842-9214-f05fc176b877 \\
        147 & 77b8ab4d-994f-43ac-8930-8ca087d7c4b4 & 148 & 788b3701-3ec9-4b67-b679-418bfa726c22 \\
        149 & 7a4e4bc8-922c-4c84-865c-25ba34136be1 & 150 & 7a5a7856-f1b6-42a4-ade9-1ca81ca0f263 \\
        151 & 7ae48c60-f143-4119-b659-15b8f485eb9a & 152 & 7aeae0e2-70ee-4705-821d-1bba5d5b2ddd \\
        153 & 7b1e1ff9-bb85-49be-b01d-d6424be18cd0 & 154 & 7b6c7e24-c58a-49fc-a5bb-d57b80e5b4c3 \\
        155 & 7b7617bd-57cc-468e-9c91-40c4ec2bcb3d & 156 & 7c4cc09e-7a92-40dd-8338-b2286535c4ed \\
        157 & 7e287123-70ca-47b9-8521-47db09b69b14 & 158 & 7e429b8d-a3f0-4ed0-9b58-08957d00b127 \\
        159 & 7f52cab9-535c-4835-ac8c-391ee64dc930 & 160 & 7ff48d5b-2df2-49da-b500-a5150ffc7f18 \\
        161 & 81c425f5-78f3-4771-afd6-3d2973825947 & 162 & 82279c77-8fc6-46f6-9622-3ba96f61b477 \\
        163 & 82bc8d6a-36eb-4d2d-8801-ef714fb1e55a & 164 & 82e3c869-49f6-4305-a7ce-f3e64a0618e7 \\
        165 & 841b50aa-df53-47bd-a73a-22d3a9f73160 & 166 & 8472fece-c7dd-4241-8d65-9b3cd1a0b568 \\
        167 & 873cafdd-a581-47f6-8b33-b9696ddb7b05 & 168 & 881deb30-9549-4583-a841-8270c65f2a17 \\
        169 & 88fe4b2d-3040-4c70-9a70-546a47764b48 & 170 & 8979838c-54a5-4454-a2b8-3d135a1a5c8f \\
        171 & 8ba5ae7a-5ae5-4eab-9fcc-5dd4fe3abf89 & 172 & 8f080098-ddb1-424c-b438-4e96e5e4786e \\
        173 & 91190194-f406-4cd6-b3f9-c43fac942b22 & 174 & 9195653c-f4aa-453d-aa95-787f6ccfaae9 \\
        175 & 930fdb3b-11a8-46fe-9bac-577332e2640e & 176 & 936321ce-5236-426a-9a20-e0e3c5dc536f \\
        177 & 937087b6-f668-4ba6-9110-60682ee33441 & 178 & 9439a27b-18ae-42d8-9778-5f68f891805e \\
        179 & 94d95f96-9699-4208-98ba-3c3119edf9c2 & 180 & 9656a811-9b5b-4ddf-99c7-5117bcef0626 \\
        181 & 971cbb5b-3cbf-4ff7-9e24-b5c84fcebfa6 & 182 & 982d12a5-beab-424f-8d38-d2a48429e511 \\
        183 & 99146c54-4f37-4ab8-9327-5f3291665e1e & 184 & 9b7bc335-06b5-4cd3-9119-1a649c478509 \\
        185 & 9bc3cc16-074a-45ac-9bdc-b2a362e1daf3 & 186 & 9cf05d24-6bd9-4dae-8967-f67d88f5d38a \\
        187 & 9d425400-e9b2-4424-9a4b-d4c7abac4140 & 188 & 9ec204e4-f0a3-42f8-8458-b772a6797cab \\
        189 & 9f3bb592-209d-43bc-bb47-d77d9df56504 & 190 & 9f935cce-0a9f-435f-8007-817732bfc0a5 \\
        191 & a01fbce3-2793-461f-ab86-43680ccbae25 & 192 & a097acff-6266-4291-9fbd-137af7ecd439 \\
        \bottomrule
    \end{tabular}
\end{table}

\clearpage
\begin{table}[H]
    \centering
    \small
    \ContinuedFloat
    \caption{Complete task IDs for the 286-task OSWorld-Verified subset (continued).}
    \renewcommand{\arraystretch}{1.07}
    \begin{tabular}{@{}r@{\hspace{5pt}}>{\ttfamily}l@{\hspace{16pt}}r@{\hspace{5pt}}>{\ttfamily}l@{}}
        \toprule
        No. & \multicolumn{1}{l}{Task ID} & No. & \multicolumn{1}{l}{Task ID} \\
        \midrule
        193 & a10b69e1-6034-4a2b-93e1-571d45194f75 & 194 & a1af9f1c-50d5-4bc3-a51e-4d9b425ff638 \\
        195 & a434992a-89df-4577-925c-0c58b747f0f4 & 196 & a462a795-fdc7-4b23-b689-e8b6df786b78 \\
        197 & a4d98375-215b-4a4d-aee9-3d4370fccc41 & 198 & a503b07f-9119-456b-b75d-f5146737d24f \\
        199 & a53f80cd-4a90-4490-8310-097b011433f6 & 200 & a669ef01-ded5-4099-9ea9-25e99b569840 \\
        201 & a728a36e-8bf1-4bb6-9a03-ef039a5233f0 & 202 & a74b607e-6bb5-4ea8-8a7c-5d97c7bbcd2a \\
        203 & a82b78bb-7fde-4cb3-94a4-035baf10bcf0 & 204 & a96b564e-dbe9-42c3-9ccf-b4498073438a \\
        205 & a9f325aa-8c05-4e4f-8341-9e4358565f4f & 206 & aa3a8974-2e85-438b-b29e-a64df44deb4b \\
        207 & aa4b5023-aef6-4ed9-bdc9-705f59ab9ad6 & 208 & aad10cd7-9337-4b62-b704-a857848cedf2 \\
        209 & abed40dc-063f-4598-8ba5-9fe749c0615d & 210 & ac1b39ff-ee4d-4483-abce-c117e98942f0 \\
        211 & acb0f96b-e27c-44d8-b55f-7cb76609dfcd & 212 & aceb0368-56b8-4073-b70e-3dc9aee184e0 \\
        213 & adf5e2c3-64c7-4644-b7b6-d2f0167927e7 & 214 & af23762e-2bfd-4a1d-aada-20fa8de9ce07 \\
        215 & af630914-714e-4a24-a7bb-f9af687d3b91 & 216 & b070486d-e161-459b-aa2b-ef442d973b92 \\
        217 & b148e375-fe0b-4bec-90e7-38632b0d73c2 & 218 & b337d106-053f-4d37-8da0-7f9c4043a66b \\
        219 & b3d4a89c-53f2-4d6b-8b6a-541fb5d205fa & 220 & b4f95342-463e-4179-8c3f-193cd7241fb2 \\
        221 & b5062e3e-641c-4e3a-907b-ac864d2e7652 & 222 & b6781586-6346-41cd-935a-a6b1487918fc \\
        223 & b7895e80-f4d1-4648-bee0-4eb45a6f1fa8 & 224 & bb5e4c0d-f964-439c-97b6-bdb9747de3f4 \\
        225 & bb83cab4-e5c7-42c7-a67b-e46068032b86 & 226 & bba3381f-b5eb-4439-bd9e-80c22218d5a7 \\
        227 & bc2b57f3-686d-4ec9-87ce-edf850b7e442 & 228 & bf4e9888-f10f-47af-8dba-76413038b73c \\
        229 & c1fa57f3-c3db-4596-8f09-020701085416 & 230 & c288e301-e626-4b98-a1ab-159dcb162af5 \\
        231 & c59742c0-4323-4b9d-8a02-723c251deaa0 & 232 & c6bf789c-ba3a-4209-971d-b63abf0ab733 \\
        233 & c7c1e4c3-9e92-4eba-a4b8-689953975ea4 & 234 & c82632a4-56b6-4db4-9dd1-3820ee3388e4 \\
        235 & c867c42d-a52d-4a24-8ae3-f75d256b5618 & 236 & ce2b64a2-ddc1-4f91-8c7d-a88be7121aac \\
        237 & ce88f674-ab7a-43da-9201-468d38539e4a & 238 & d1acdb87-bb67-4f30-84aa-990e56a09c92 \\
        239 & d38192b0-17dc-4e1d-99c3-786d0117de77 & 240 & d52d6308-ec58-42b7-a2c9-de80e4837b2b \\
        241 & d53ff5ee-3b1a-431e-b2be-30ed2673079b & 242 & d681960f-7bc3-4286-9913-a8812ba3261a \\
        243 & d9b7c649-c975-4f53-88f5-940b29c47247 & 244 & da46d875-6b82-4681-9284-653b0c7ae241 \\
        245 & da52d699-e8d2-4dc5-9191-a2199e0b6a9b & 246 & da922383-bfa4-4cd3-bbad-6bebab3d7742 \\
        247 & dcbe20e8-647f-4f1d-8696-f1c5bbb570e3 & 248 & dd60633f-2c72-42ba-8547-6f2c8cb0fdb0 \\
        249 & dd84e895-72fd-4023-a336-97689ded257c & 250 & df67aebb-fb3a-44fd-b75b-51b6012df509 \\
        251 & dfac9ee8-9bc4-4cdc-b465-4a4bfcd2f397 & 252 & e0df059f-28a6-4169-924f-b9623e7184cc \\
        253 & e135df7c-7687-4ac0-a5f0-76b74438b53e & 254 & e1e75309-3ddb-4d09-92ec-de869c928143 \\
        255 & e246f6d8-78d7-44ac-b668-fcf47946cb50 & 256 & e2b5e914-ffe1-44d2-8e92-58f8c5d92bb2 \\
        257 & e2dd0213-26db-4349-abe5-d5667bfd725c & 258 & e528b65e-1107-4b8c-8988-490e4fece599 \\
        259 & ea98c5d7-3cf9-4f9b-8ad3-366b58e0fcae & 260 & eabc805a-bfcf-4460-b250-ac92135819f6 \\
        261 & eb03d19a-b88d-4de4-8a64-ca0ac66f426b & 262 & eb303e01-261e-4972-8c07-c9b4e7a4922a \\
        263 & ec4e3f68-9ea4-4c18-a5c9-69f89d1178b3 & 264 & ec71221e-ac43-46f9-89b8-ee7d80f7e1c5 \\
        265 & ecb0df7a-4e8d-4a03-b162-053391d3afaf & 266 & ecc2413d-8a48-416e-a3a2-d30106ca36cb \\
        267 & ed43c15f-00cb-4054-9c95-62c880865d68 & 268 & ee9a3c83-f437-4879-8918-be5efbb9fac7 \\
        269 & ef9d12bd-bcee-4ba0-a40e-918400f43ddf & 270 & f0b971a1-6831-4b9b-a50e-22a6e47f45ba \\
        271 & f178a4a9-d090-4b56-bc4c-4b72a61a035d & 272 & f201fbc3-44e6-46fc-bcaa-432f9815454c \\
        273 & f23acfd2-c485-4b7c-a1e7-d4303ddfe864 & 274 & f3977615-2b45-4ac5-8bba-80c17dbe2a37 \\
        275 & f4aec372-4fb0-4df5-a52b-79e0e2a5d6ce & 276 & f5d96daf-83a8-4c86-9686-bada31fc66ab \\
        277 & f723c744-e62c-4ae6-98d1-750d3cd7d79d & 278 & f79439ad-3ee8-4f99-a518-0eb60e5652b0 \\
        279 & f7dfbef3-7697-431c-883a-db8583a4e4f9 & 280 & f8cfa149-d1c1-4215-8dac-4a0932bad3c2 \\
        281 & f9584479-3d0d-4c79-affa-9ad7afdd8850 & 282 & f9be0997-4b7c-45c5-b05c-4612b44a6118 \\
        283 & fba2c100-79e8-42df-ae74-b592418d54f4 & 284 & fbb548ca-c2a6-4601-9204-e39a2efc507b \\
        285 & fc6d8143-9452-4171-9459-7f515143419a & 286 & fe41f596-a71b-4c2f-9b2f-9dcd40b568c3 \\
        \bottomrule
    \end{tabular}
\end{table}

\end{document}